\documentclass[journal]{IEEEtran}
\pdfoutput=1

\usepackage{times}
\usepackage{epsfig}
\usepackage{graphicx}
\usepackage{wrapfig}
\usepackage{amsmath}
\usepackage{amssymb}
\usepackage{expl3}
\usepackage{multirow}
\usepackage{tabularx}
\usepackage{algorithm}
\usepackage{algpseudocode}
\usepackage{flushend}
\usepackage{rotating}

\ExplSyntaxOn
\newcommand\latinabbrev[1]{
  \peek_meaning:NTF . {
    #1\@}%
  { \peek_catcode:NTF a {
      #1.\@ }%
    {#1.\@}}}
\ExplSyntaxOff

\DeclareMathOperator*{\argmax}{arg\,max}

\def\eg{\latinabbrev{\emph{e.g}}}
\def\ie{\latinabbrev{\emph{i.e}}}

\usepackage{hyperref}

\begin{document}

\title{Who did What at Where and When: Simultaneous Multi-Person Tracking and Activity Recognition}

\author{Wenbo Li, Ming-Ching Chang, Siwei Lyu\\
\and
Computer Science Department, University at Albany, State University of New York\\
\and
$\{$wli20,mchang2,slyu$\}$@albany.edu \\
{\footnotesize  Ming-Ching Chang is the corresponding author.}
}


\maketitle

\begin{abstract}
We present a bootstrapping framework to simultaneously improve multi-person tracking and activity recognition at individual, interaction and social group activity levels. The inference consists of identifying trajectories of all pedestrian actors, individual activities, pairwise interactions, and collective activities, given the observed pedestrian detections. Our method uses a graphical model to represent and solve the joint tracking and recognition problems via multi-stages: (1) activity-aware tracking, (2) joint interaction recognition and occlusion recovery, and (3) collective activity recognition. We solve the {\em where} and {\em when} problem with visual tracking, as well as the {\em who} and {\em what} problem with recognition. High-order correlations among the visible and occluded individuals, pairwise interactions, groups, and activities are then solved using a hypergraph formulation within the Bayesian framework. Experiments on several benchmarks show the advantages of our approach over state-of-art methods.
\end{abstract}

\IEEEpeerreviewmaketitle
\section{Introduction}
\label{sec_1}


Multi-person activity recognition is a major component of many applications, \eg, video surveillance and traffic control. The problem entails the inference of the activities, the actors and their motion trajectories, as well as the time dynamics of the events. This task is challenging, since the activities are analyzed from both the spontaneous individual actions and the complex social dynamics involving groups and crowds \cite{Social:Signal:Processing}. We aim to address not only the {\bf where} and {\bf when} problem by visual tracking, but also the {\bf who} and {\bf what} problem by activity recognition.

While advanced methods for person detection are becoming more reliable \cite{Unified:Multi:Scale:Detection,POI}, most existing activity recognition approaches rely on visual tracking following a tracking-by-detection paradigm. These methods either fail to consider social interactions while inferring activities \cite{Hierarchical:Deep:Temporal:Model,Combining:Per-frame:and:Per-track:Cues,Flow:Model} or have difficulties recognizing the structural correlations of actions and interactions \cite{Choi:Savarese:Tracking:Collective:Activity:Reog:ECCV2012,Choi:Collective:Activities:PAMI2014,Structure:Inference:Machines}. In particular, there are two major challenges: \emph{(i)} ineffective tracking due to frequent occlusions in groups and crowds, and \emph{(ii)} the lack of a suitable methodology to infer the complex but salient structures involving social dynamics and groups.

\begin{figure}[t]
\centerline{
  \includegraphics[width=\linewidth]{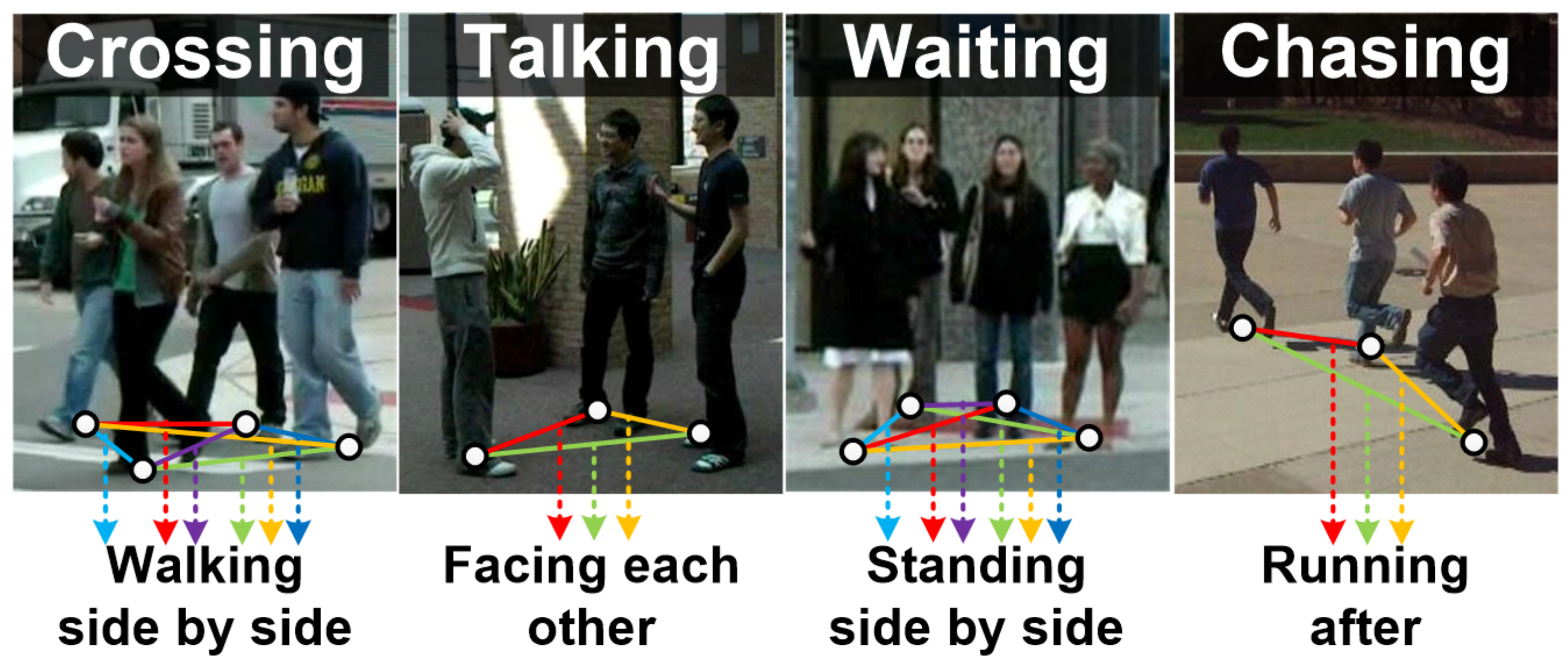}
  \vspace{0.1cm}
}
\centerline{
  \includegraphics[width=0.4\linewidth]{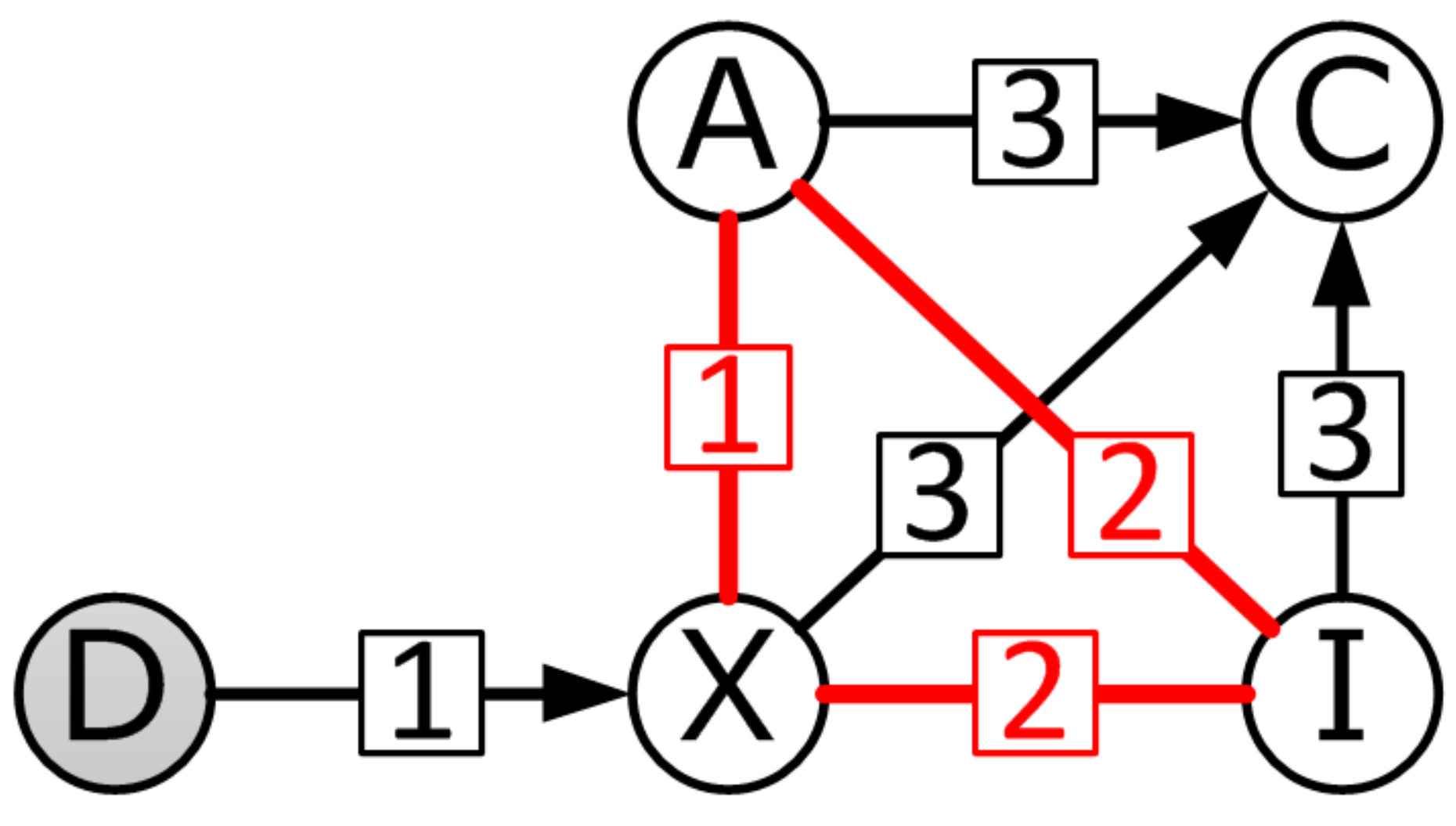}
  \vspace{-0.2cm}
}
\caption{\em
(Top)
This work is based on two main hypotheses that:
{\em (i)} Multi-person tracking and activity recognition can be jointly solved using an improved, unified framework.
{\em (ii)}
Group {\em collective activities} (crossing, talking, waiting, chasing, {\em etc.}) can be characterized by the cohesive {\em pairwise interactions} (walking side by side, facing each other, standing side by side, running after, {\em etc.}, see Table \ref{tab:interaction:formulas}) within the group. See $\S$ \ref{sec:formulation}.
(Bottom)
The dependency graph that can jointly infer the target tracking ($X$), individual activities ($A$), pairwise interactions ($I$), and collective activities ($C$), all from the input detections ($D$). Numbers on the edges indicate the inference stages in the multi-stage updating scheme.
}
\label{fig:overview}
\end{figure}

In this paper, we address both challenges using a bootstrapping framework to simultaneously improve the two tasks of multi-person tracking and social group activity recognition. We take state-of-the-art person detections \cite{Unified:Multi:Scale:Detection,POI} as input to perform initial multi-person tracking. We then recognize stable group structures including the temporally cohesive {\em individual activities} (such as walking) and {\em pairwise interactions} (such as walking side-by-side, see Fig.\ref{fig:overview} to robustly infer collective social activities (such as street crossing in a group) in multiple stages. Auxiliary inputs such as body orientation detections can be considered within the stages as well. The recognized activities and salient grouping structures are used as priors to recover occluded detections and false associations to improve performance.

We explicitly explore the correlations of {\em pairwise interactions} (of two individuals) and their {\em group activities} (within the group of more individuals) during the optimization.
Observe in Fig.\ref{fig:overview} that group activities generally consist of correlated pairwise interactions, which we have exploited in the multi-stage inference steps. In our method, the multi-target tracking and the recognition of individual/group activities are jointly optimized, such that consistent activity labels characterizing the dynamics of the individuals and groups can be obtained. The individual and group activities are formulated using a dynamic graphical model, and high-order correlations are represented using {\bf hypergraphs}. The simultaneous pedestrian tracking and multi-person activity recognition problems are then to be solved using an efficient cohesive cluster search in the hypergraphs.

The main contribution of this work is two-fold. First, we propose a new framework that can jointly solve the two tasks of real-time simultaneous tracking and activity recognition. Explicit modeling of the correlations among the individual activities, pairwise interactions, and collective activities leads to a consistent solution. Second, we propose a hypergraph formulation to infer the high-order correlations among social dynamics, occlusions, groups, and activities in multi-stages. Simultaneous tracking and activity recognition are formulated as a bootstrapping framework, which can be solved efficiently using the search of cohesive clusters in the hypergraphs. This hypergraph solution is general that it can be extended to include additional scenarios or constraints in new applications.

Experiments on several benchmarks show the advantages of our method with improvements in both activity recognition and multi-person tracking.
Our method is easily deployable to real-world applications, since: \emph{(i)} camera calibration is not required; \emph{(ii)} {\em online} video streams can be processed by considering a time window in a round; \emph{(iii)} the computation can be performed in real-time (about 20 FPS, not including the input detection steps).

\section{Related Works}
\label{sec_2}

There exists a tremendous amount of multi-person tracking and activity recognition works. See \cite{Human:Activity:Analysis,Multiple:Object:Tracking:A:Review} for survey. Our work is most related to the {\em collective activity} recognition, which are organized into the following three categories --- recognition based on {\em (i)} detection, {\em (ii)} tracking, and {\em (iii)} simultaneous tracking and recognition.

\subsection{Collective Activity Recognition based on Detection}

A hierarchical model is used in \cite{Discriminative:Latent:Models} to recognize collective activities by considering the person-person and group-person contextual information. The work of \cite{Deep:Structured:Models} uses hierarchical deep neural networks with a graphical model to recognize collective activities based on the dependencies of individual activities. This work is further extended in \cite{Structure:Inference:Machines}, where the individual and collective activities are iteratively recognized using RNN with refinements. Multi-instance learning is used in \cite{Visual:recognition:by:counting:instances} to recognize collective activities by inferring the cardinality relations of individual activities. A recurrent CNN is used in \cite{Social:Scene:Understanding} for the joint target detection and activity recognition.

\subsection{Collective Activity Recognition based on Tracking}

In this category, individual target trajectories are used as the input to recognize collective activities. Collective activities are recognized in \cite{Augmented:CAD:umich} using random forests for the spatio-temporal volume classification. A two-stage deep temporal neural network is used in \cite{Hierarchical:Deep:Temporal:Model}, where the first stage recognizes individual activities, and the second stage aggregates individual observations to recognize  collective activities. In \cite{Learning:Latent:Constituents}, the key {\em constituents} of activities and their relationships are used to recognize collective activities. A graphical model is developed in \cite{HiRF:Group:Activity:ECCV2014} to capture high-order temporal dependencies of video features. The {\em and-or graph} \cite{AndOrGraph:Scheduling:Activity:ICCV2013} is applied for video parsing and activity querying, where the detectors and trackers are launched upon receiving queries. A RNN architecture is designed in \cite{Recurrent:Modeling:Interaction:Context} to model high-order social group interaction contexts.

\subsection{Simultaneous Tracking and Activity Recognition}

Only very few works deal with the problem of simultaneous multi-person tracking and activity recognition. In \cite{Combining:Per-frame:and:Per-track:Cues}, per-frame and per-track cues extracted from an appearance-based tracker are combined to capture the regularity of individual actions. A network flow-based model is used in \cite{Flow:Model} to link detections while inferring collective activities. However, these two methods did not consider pairwise interactions for activity recognition. In \cite{Choi:Savarese:Tracking:Collective:Activity:Reog:ECCV2012,Choi:Collective:Activities:PAMI2014}, the tracking and activity recognition are formulated as a joint energy maximization problem, which is solved by belief propagation with branch-and-bound. However high-order correlations among individual and pairwise activities are not considered, which limits the activity recognition performance. 
\section{Method}
\label{sec_3}

We start with notation definition in our method. Given an input video sequence, consider the most recent time window $T = [t-\tau, t]$ in an {\em online} fashion, and denote previous time frames $[1, t-\tau-1]$ as $T^{\prime}$. Let $D_{T}$
represent a set of target {\em detections} obtained using person detectors {\em e.g.} \cite{Unified:Multi:Scale:Detection,POI}. Let $X_{T^{\prime}}$ represent the set of existing target trajectories. Let $A_{T^{\prime}}$, $I_{T^{\prime}}$, and $C_{T^{\prime}}$ represent the set of recognized individual activities, pairwise interactions, and collective activities, respectively. Given $D_{T}$, our approach aims to simultaneously solve the multi-person tracking and activity recognition problems, by inferring the following four terms within $T$:
{\em (i)} {\em target trajectories}
$X_{T} = \{ \texttt{x}_1, \ldots, \texttt{x}_b \}$, where $b$ is the number of observed targets,
{\em (ii)} {\em individual activity} labels
$A_{T} = \{ \texttt{a}_1, \ldots, \texttt{a}_b \}$,
{\em (iii)} {\em pairwise interaction} labels $I_{T} = \{ \texttt{i}_{1,2}, \texttt{i}_{1,3}, \ldots, \texttt{i}_{2,3}, \ldots, \texttt{i}_{b-1,b} \}$,
and {\em (iv)} {\em collective activity} labels $C_{T} = \{ \texttt{c}_{t-\tau}, \ldots, \texttt{c}_{t} \}$, where $\texttt{c}_f$ represents the collective activity with the most involved targets in the $f$-th frame.
After a time window is processed, the method will extend target tracklets, update activity labels, and move on to the next time window: $X_{1:t} = [X_{T^{\prime}}, X_{T}]$, $A_{1:t} = [A_{T^{\prime}}, A_{T}]$, $I_{1:t} = [I_{T^{\prime}}, I_{T}]$, and $C_{1:t} = [C_{T^{\prime}}, C_{T}]$.
To simplify notions, we omit the temporal indices to represent the variables within $[t-\tau, t]$ as $X,A,I,C$, and represent previous variables as $X^{\prime},A^{\prime},I^{\prime},C^{\prime}$, \ie $X' = X_{T'}$, $A' = A_{T'}$, $I' = I_{T'}$, $C' = C_{T'}$.

\begin{table*}[t]
\caption{\em
Notations for video activities, problem formulation and visual tracking.
\vspace{-0.3cm}
}
\centering
\label{tab:symbols:1}

\begin{tabular}{|c|c|l|}
\hline
  & symbol & description \\
\hline\hline


\multirow{15}{0.2cm}{ \begin{sideways} Video Activities \end{sideways} }
  & $\texttt{x}$ & a target trajectory \\
\cline{2-3}
  & $\texttt{a}$ & an individual activity label, {\em e.g.} $\in \{ standing, walking, running \}$ \\
\cline{2-3}
  & $\texttt{i}$ & a pairwise interaction label, {\em e.g.} {\em approaching} (AP), {\em facing-each-other} (FE), {\em standing-in-a-row} (SR), ... \\
\cline{2-3}
  & $\texttt{c}$ & a collective activity label, {\em e.g.}, {\sc crossing}, {\sc walking}, {\sc gathering}, ... \\
\cline{2-3}
  & $b$ & number of observed targets (tracklets) \\
\cline{2-3}
\\[-1em]
  & $T$ & video time window of length $\tau$ prior to time $t$, {\em i.e.}, $T = [t-\tau, t]$ \\
\cline{2-3}
  & $D$ & person detections (bounding boxes) \\
\cline{2-3}
  & $X$ & target trajectories, $X_{T} = \{ \texttt{x}_1, \ldots, \texttt{x}_b \}$ \\
\cline{2-3}
  & $A$ & individual activity classes, $A_{T} = \{ \texttt{a}_1, \ldots, \texttt{a}_b \}$ \\
\cline{2-3}
  & $I$ & pairwise interaction classes, $I_{T} = \{ \texttt{i}_{1,2}, \texttt{i}_{1,3}, \ldots, \texttt{i}_{2,3}, \ldots, \texttt{i}_{b-1,b} \}$ \\
\cline{2-3}
  & $C$ & collective activity classes, $C_{T} = \{ \texttt{c}_{t-\tau}, \ldots, \texttt{c}_{t} \}$ \\
\cline{2-3}
\\[-1em]
  & $T', X', A', I', C'$ & existing entities prior to time window $T$, $X' = X_{T'}$, $A' = A_{T'}$, $I' = I_{T'}$, $C' = C_{T'}$ \\
\cline{2-3}
  & $n_A$ & number of individual activity classes, $n_A = 2$ in the CAD and Augmented-CAD datasets, $n_A = 3$ in the New-CAD dataset \\
\cline{2-3}
 &\multirow{2}{*}{$n_I$}
  & number of interaction classes, which is also the number of sub-hypergraphs used in our \\
  & &  method, $n_I = 8$ in the CAD and Augmented-CAD datasets, $n_I = 9$ in the New-CAD dataset \\
\cline{2-3}
  & $n_C$ & number of collective activity classes, $n_C = 5$ in CAD, $n_C = 6$ in Augmented-CAD, $n_C = 6$ in New-CAD datasets \\

\hline


\multirow{11}{0.2cm}{ \begin{sideways} Problem Formulation \end{sideways} }
  & $\mathbf{Pr}$ & a joint distribution \\
\cline{2-3}
  & $f_1, f_2, f_3$ & confidence terms from the decomposition of $\mathbf{Pr}$ \\
\cline{2-3}
  & $\varphi_1, \varphi_2, \varphi_3$ & clique potential functions in the Markov random field \\
\cline{2-3}
\\[-1em]
  & $X^*, A^*, I^*, C^*$ & updated terms of $X, A, I, C$ after an optimization stage, respectively \\
\cline{2-3}
\\[-1em]
  & $X^{\ddagger}, A^{\ddagger}$ & updated terms of $X^*, A^*$ after an optimization stage, respectively \\
\cline{2-3}
  & $p_{ds}$ & the {\em distance} likelihood term for estimating the interaction between two targets $$\\
\cline{2-3}
  & $p_{gc}$ & the {\em group connectivity} term for estimating the interaction between two targets \\
\cline{2-3}
  & $p_{aa}$ & the {\em individual activity agreement} term for estimating the interaction between two targets \\
\cline{2-3}
  & $p_{dc}$ & the {\em distance change type} likelihood term for estimating the interaction between two targets \\
\cline{2-3}
  & $p_{dr}$ & the {\em facing direction} likelihood term for estimating the interaction between two targets \\
\cline{2-3}
  & $p_{fs}$ & the {\em frontness/sideness} likelihood term for estimating the interaction between two targets \\

\hline


\multirow{8}{0.2cm}{ \begin{sideways} Tracking \end{sideways} }
  & $\bar{\texttt{x}}$ & a candidate tracklet \\
\cline{2-3}
\\[-1em]
  & $\bar{X}$ & the set of all candidate tracklets \\
\cline{2-3}
  & $\bar{\texttt{a}}$ & a (putative) individual activity of a candidate tracklet \\
\cline{2-3}
\\[-1em]
  & $\bar{A}$ & the set of (putative) individual activities for all candidate tracklets \\
\cline{2-3}
  & $\theta_a$ & the {\em appearance} similarity for tracklet linking \\
\cline{2-3}
  & $\tau_a$ & time threshold for appearance-based tracklet linking \\
\cline{2-3}
  & $\oplus$ & operator $\oplus$ represents the association of two tracklets \\
\cline{2-3}
  & $h$ & the number of hypothetical tracklets to generate from an existing tracklet $\texttt{x}_i'$, $h = 9$ \\

\hline

\end{tabular}
\end{table*}

\begin{table*}[t]
\caption{ \em
Graph and hypergraph notations.
\vspace{-0.3cm}
}
\centering
\label{tab:symbols:2}

\begin{tabular}{|c|c|l|}
\hline
  & symbol & description \\
\hline\hline


\multirow{21}{0.5cm}{ \begin{sideways} Hypergraph \end{sideways} }
  & $\cal H$ & hypergraph ${\cal H} = (V, E, W)$ \\
\cline{2-3}
  & $\cal H_T$ & tracking hypergraph ${\cal H_T} = (V_{\cal T}, E_{\cal T}, W_{\cal T})$\\
\cline{2-3}
  & $\cal H_R$ & activity recognition hypergraph ${\cal H_R} = (V_{\cal R}, E_{\cal R}, W_{\cal R})$\\
\cline{2-3}
  & $V$ & the vertex set of a hypergraph\\
\cline{2-3}
  & $E$ & the hyperedge set of a hypergraph \\
\cline{2-3}
  & $W$ & the hyperedge weights of a hypergraph \\
\cline{2-3}
  & $W_a$ & the {\em appearance} hyperedge weight, working with control parameter $\lambda_a = 30$ \\
\cline{2-3}
  & $W_d$ & the {\em facing-direction} hyperedge weight, working with control parameter $\lambda_d = 1$ \\
\cline{2-3}
  & $W_g$ & the {\em geometric similarity} hyperedge weight, working with control parameter $\lambda_g = 0.5$ \\
\cline{2-3}
  & $m$ & the hyperedge degree, {\em i.e.}, the number of incident vertices of the hyperedge \\
\cline{2-3}
\\[-1em]
  & $\mathbf{e}^m$ & a $m$-degree hyperedge, $\mathbf{e}^{m} = \{ v_1^{\mathbf{e}}, \ldots, v_m^{\mathbf{e}} \}$ \\
\cline{2-3}
  & $\cal C$ & a hyperedge cluster, which is a vertex set with interconnected hyperedges \\
\cline{2-3}
\\[-1em]
  & $\kappa$ & number of vertes in a hypergraph cluster $\cal C$, $\kappa = | {\cal C} |$ \\
\cline{2-3}
\\[-1em]
  & $E^{\cal C}$ & the set of all incident hyperedges of a cluster $\cal C$ \\
\cline{2-3}
  & $\Psi$ & weighting function operated on a hypergraph cluster $\cal C$ \\
\cline{2-3}
  & $\mathbf{y}$ & the indicator vector to denote the vertex selection from $V \in {\cal H}$ to be included in ${\cal C}$  \\
\cline{2-3}
\\[-1em]
  & $\epsilon$ & $\epsilon = \frac{1}{\kappa}$ used in weight normalization \\
\cline{2-3}
\\[-1em]
  & $\delta$ & $\delta_p = \frac{y_p}{\kappa}$ used in weight normalization \\
\cline{2-3}
\\[-1em]
  & $\mathbf{p}_{ij}$ & image coordinate vector between two positions at $i$ and $j$ \\
\cline{2-3}
\\[-1em]
  & $\cal \breve{H}$ & a sub-hypergraph indexed by $\beta$, {\em i.e.}, ${\cal \breve{H}}_{\beta}$ \\
\cline{2-3}
\\[-1em]
  & $\breve{E}_{\beta}$ & the hyperedges of the sub-hypergraph ${\cal \breve{H}}_{\beta}$ corresponding to the $\beta$-th interaction class \\
\cline{2-3}
\\[-1em]
  & $\breve{W}_{\beta}$ & the hyperedge weights of the sub-hypergraph ${\cal \breve{H}}_{\beta}$ corresponding to the $\beta$-th interaction class \\

\hline


\multirow{10}{0.5cm}{ \begin{sideways} Graph \end{sideways} }
\\[-1em]
  & $\cal \tilde{G}$ & graph ${\cal \tilde{G}} = (\tilde{V}, \tilde{E}, \tilde{W})$ \\
\cline{2-3}
\\[-1em]
  & $\tilde{V}$ & the vertex set of a graph; $\tilde{V}$ is associated with $X'$ in this paper  \\
\cline{2-3}
\\[-1em]
  & $\tilde{E}$ & the edge set of a graph \\
\cline{2-3}
\\[-1em]
  & $\tilde{W}$ & the edge weights of a graph \\
\cline{2-3}
  & $e_{ij}$ & a graph edge connecting two vertices $v_i$ and $v_j$ \\
\cline{2-3}
\\[-1em]
  & $p_{corr}$ & the correlation between the activities of two targets $\texttt{x}_i$ and $\texttt{x}_j$ used to calculate weight $\tilde{W}(e_{ij})$ \\
\cline{2-3}
  & $g$ & a function to calculate the correlation between the activities of two targets \\
\cline{2-3}
  & $d$ & Eucludean distance between two targets in the image coordinate. \\
\cline{2-3}
  & $\phi_{ij}$ & the angle between the facing direction of $\texttt{x}_i$ and the relative vector from $\texttt{x}_i$ to $\texttt{x}_j$. \\
\cline{2-3}
\\[-1em]
  & ${\cal \tilde{G}}_s$ & sparse graph by discarding edges with small weights from $\cal \tilde{G}$ \\
\hline


\multirow{6}{0.5cm}{ \begin{sideways} Indices \end{sideways} }
  & $t$, $\tau$, $f$ & video frame indices \\
\cline{2-3}
\\[-1em]
  & $i$, $j$, $k$, $l$ & target tracklet indices \\
\cline{2-3}
\\[-1em]
  & $p$, $q$, $r$ & hypergraph vertex indices \\
\cline{2-3}
\\[-1em]
  & $\alpha$ & the index for hypergraph clusters {\em e.g.} $\cal C_{\alpha}$, $\cal C^{\cal T}_{\alpha}$ from $\cal H_T$ \\
\cline{2-3}
\\[-1em]
  & $\beta$ & the index for interaction classes {\em e.g.} $I_{\beta}$; $\beta$ is also the index for sub-hypergraphs {\em e.g.} ${\cal \breve{H}}_{\beta}$ \\
\cline{2-3}
\\[-1em]
  & $c$ & the index for collective activity classes $C$ \\

\hline

\end{tabular}
\end{table*}

\subsection{Problem Formulation}
\label{sec:formulation}

We aim to infer accurate trajectories of all targets ($X$) and their individual activities ($A$), pairwise interactions ($I$) and collective activities ($C$) from the observed detections ($D$).
Relationship between these variables can be expressed as the joint distribution $\textbf{Pr}(X,A,I,C|D)$ as a dependency graph in Fig.\ref{fig:overview}. Based on the conditional independence assumption of $X,A,I,C$ in the graphical model, $\textbf{Pr}(X,A,I,C|D)$ can be decomposed into three terms:
\begin{equation}
\label{eq:joint:distribution}
f_{1}(X,D) \cdot f_{2}(X,A,I) \cdot f_{3}(X,A,I,C).
\end{equation}
{\em (i)} $f_{1}(X,D)$ is the confidence of target tracking, where the calculation will be given in $\S$~\ref{sec:stage1}.
{\em (ii)} $f_{2}(X,A,I)$ models the inter-dependencies among target trajectories, individual activity and pairwise interaction labels, which is further expressed as a Markov random field (red cycle in Fig.~\ref{fig:overview}):
%
\begin{equation}
\label{eq:markov}
f_{2}(X,A,I) \;\; \sim \;\; \varphi_1(X,A) \cdot \varphi_2(A,I) \cdot \varphi_3(I,X),
\end{equation}
where $\varphi_1$, $\varphi_2$ and $\varphi_3$ are three {\em clique potential} functions capturing the inter-correlations between each variable pair. Derivation of these clique potentials will be given in $\S$~\ref{sec:stage1} and $\S$~\ref{sec:stage2}.
{\em (iii)} $f_{3}(X,A,I,C)$ reflects an important assumption that collective activities can be effectively modeled by robust inference of target trajectories, individual activities and pairwise interactions, where $\S$~\ref{sec:stage3} will provide details.

The inference of the joint tracking and recognition is then formulated as seeking $\displaystyle \argmax_{X,A,I,C} \; \log \; \textbf{Pr}(X,A,I,C|D) = $
\begin{equation}
\!\! \argmax_{X,A,I,C}
\left\{
\begin{array}{c} \!
 \log f_{1}(X,D) +  \log \varphi_1(X,A) + \log \varphi_2(A,I) \\
 + \log \varphi_3(I,X) + \log f_{3}(X,A,I,C)
\end{array}
\!
\right\}. \nonumber
\end{equation}
However, standard iterative optimization such as {\em block coordinate descent} is not practical due to that: {\em (i)} the coupling of variables $X,A,I,C$ is still complicated; {\em (ii)} each of these variables represents a superset of time-dependent variables, so their joint optimization will be very inefficient; {\em (iii)} a real-time processing method is desired.
We adopt a heuristic approximate solution using {\bf multi-stage} {\em updating scheme}, which first jointly updates $X$, $A$, and then updates $I$, followed by the update of $C$. Our strategy is based on an important hypothesis that {\em inferring pairwise interactions $I$ is crucial in resolving the entire optimization}, because $I$ is the knob governing the representations in-between $X, A$ and $C$.

Our updating scheme shares spirit with the standard Gibbs sampling and MH-MCMC method for the inference in probabilistic graphical models.
The updating scheme takes the following three stages:

\noindent \textbf{Stage 1} {\bf activity-aware tracking} ($\S$~\ref{sec:stage1}), where individual target trajectories and activity labels are updated using:
\begin{equation}
\label{eq:stage1:objective}
(X^*,A^*) = \argmax_{X,A} \; \log f_{1}(X,D) + \log \varphi_1(X,A).
\end{equation}

\noindent \textbf{Stage 2} {\bf joint interaction recognition and occlusion recovery} ($\S$~\ref{sec:stage2}), where the interaction labels together with the target trajectories and activities are updated using:
\begin{equation}
\hspace{-0.3cm}
(X^{\ddagger}, A^{\ddagger}, I^*) = \argmax_{X^*, A^*, I} \log \varphi_2(A^*,I) + \log \varphi_3(I,X^*).
\label{eq:stage2:objective}
\end{equation}

\noindent \textbf{Stage 3} {\bf collective activity recognition} ($\S$~\ref{sec:stage3}), where the collective activity labels are updated using:
\begin{equation}
\label{eq:stage3:objective}
C^* = \argmax_C \; \log f_3(X^{\ddagger},A^{\ddagger},I^*,C).
\end{equation}

We will show in $\S$~\ref{sec:stage1} and $\S$~\ref{sec:stage2} that we model {\em high-order} correlations among $X$, $A$ and $I$ using two respective {\em hypergraphs}. The clique potentials $\varphi_1, \varphi_2, \varphi_3$ in \textbf{Stage~1} and \textbf{Stage~2} can be derived as the optimization of maximal weight search over the two hypergraphs, in order to infer $X, A, I$. \textbf{Stage~3} infers $C$ using a probabilistic formulation based on the inferred $X, A, I$.

Notations for video activities, problem formulation and tracking are summarized in Table~\ref{tab:symbols:1}, where graph and hypergraph related notations are summarized in Table~\ref{tab:symbols:2}.


\subsection{Cohesive Cluster Search on the Hypergraph}
\label{sec:hypergraph:opt}

We define an {\em undirected hypergraph} ${\cal H} = (V,E,W)$, where $V=\{ v_r \}$ denotes the {\em vertex} set of ${\cal H}$, where $r$ denotes vertex index. An {\em undirected hyperedge} with $m$-incident vertices is defined as $\mathbf{e}^{m} = \{ v_1^{\mathbf{e}}, \ldots, v_m^{\mathbf{e}} \}$, where $m$ is the {\em degree} of the hyperedge. The set of all $m$-degree hyperedges is denoted as $E  = \{\mathbf{e}^{m}\}$.
The {\em weights} of hyperedges are denoted as $W: E \rightarrow \mathbb{R}$, {\em i.e.}, each hyperedge is associated with a weight.

\begin{figure*}[t]
  \centering
  \includegraphics[width=1\linewidth]{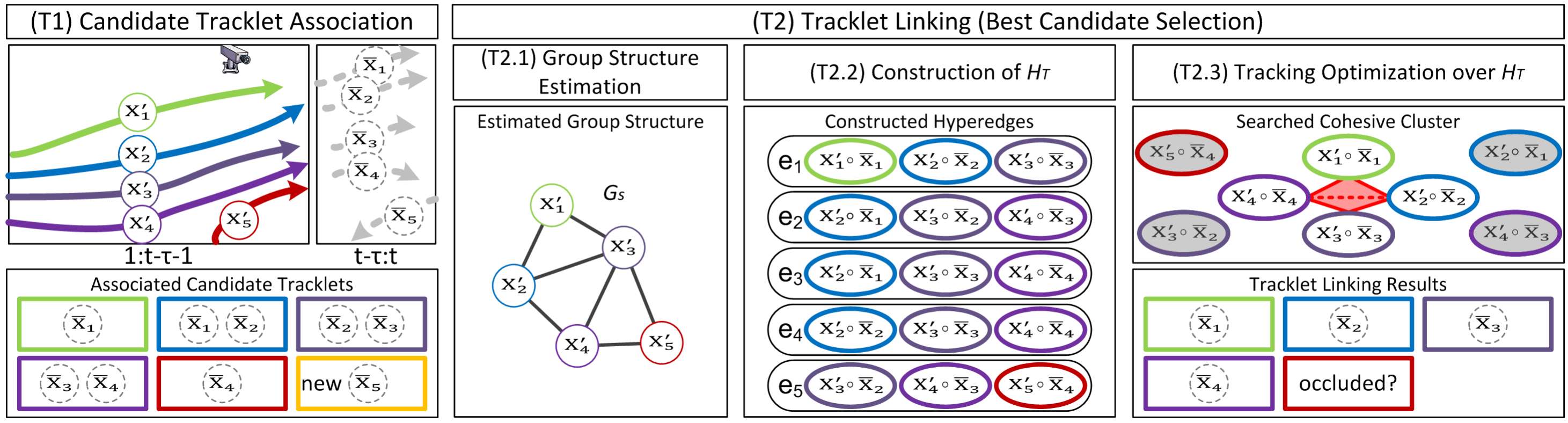}\\ 
  \caption{\em
{\bf Stage 1 activity-aware tracking.}
Given five targets $\texttt{x}^{\prime}_{1}, \ldots, \texttt{x}^{\prime}_{5}$, and new tracklets $\bar{\texttt{x}}_{1}, \ldots, \bar{\texttt{x}}_{5}$, step {\bf (T1)} optimizes the association of candidate tracklets with existing target tracks.
Step {\bf (T2)} determines the best candidate assignments for tracklet linking in three steps.
(T2.1) estimates the group structure using graph $\mathcal{\tilde{G}}_s$, where the edges represent the correlations of activities between individuals.
(T2.2) constructs hypergraph $\mathcal{H}_{T}$ with hyperedges $e_{1}, \ldots, e_{5}$ based on the estimated group structure.
(T2.3) solves the candidate tracklet linking and infers the possible occlusions in an optimization over $\mathcal{H}_\mathcal{T}$.
}
\label{fig:tracking:scheme}
\end{figure*}

We use the hypergraphs to represent both (1) the detection-tracklet association for tracking $(X', X)$, and (2) the correlations among individual activities $A$ and pairwise interactions $I$. The joint problem of multi-person tracking (with possible refinements) and group activity recognition can be solved using a standard {\em cohesive cluster search} on the hypergraph \cite{Robust:Clustering}. A {\em cluster} ${\cal C}$ within a hypergraph is a vertex set with interconnected hyperedges. We use $\kappa = \left| \mathcal{C} \right|$  to denote the number of vertices in $\cal C$, and $E^{\mathcal{C}}$ to denote the set of all incident hyperedges of ${\mathcal{C}}$. A cluster is {\em cohesive} if its vertices are interconnected by a large amount of hyperedges with dense weights. Denote $\Psi( \cdot )$ the {\bf weighting function} that measures the weight of a cluster. For a vertex $v_r \in V$, the cohesive cluster search optimization is to determine a large cluster ${\cal C}(v_r)$ with dense weights:
\begin{equation}
{\mathcal{C}(v_r)}^{*} = \argmax_{\mathcal{C}(v_r)} \; \Psi \left( \mathcal{C}(v_r) \right) ~\textrm{s.t.}~\mathcal{C}(v_r) \subset V.
\label{eq:orig:hyper:objective}
\end{equation}
We use {\em indicator vector} $\mathbf{y} = (y_1, ..., y_{|V|})$,
$y_p \in \{0,1\}$ to denote the selection of vertices from $\cal H$ to be included in $\cal C$: $y_p = 1$ for $v_p \in {\cal C}$, and $y_p = 0$ otherwise.
The selection is constrained such that up to $\kappa$ vertices including $v_r$ are enclosed in ${\cal C}$, such that $\sum_{p=1}^{|V|} y_p = \kappa$, and $y_r = 1$.

The design of $\Psi(\cdot)$ affects the resulting cluster $\mathcal{C}$ from the search. Typical $\Psi(\cdot)$ can be the total weight of all incident hyperedges. However, direct maximization of the total weight leads to a large cluster that is not necessarily cohesive. Instead, we maximize a {\em normalized} weight, which is the total weight divided by the cardinality of all incident hyperedges.
This normalization also enables continuous optimization.
For $\mathcal{C}$ with $\kappa$ vertices and $m$-degree hyperedges, this normalizer is $\kappa^m$. Our weighting function $\Psi(\mathcal{C}(v_r))$ is:
\begin{equation}
\hspace{-0.2cm}
\frac{ \sum\limits_{\mathbf{e}^{m} \in E^{\mathcal{C}}} W(\mathbf{e}^{m}) }{ \kappa^{m} } = \! \sum_{v_p,\ldots,v_q \in V}
\left( \!\!
W( \overbrace{v_p, \ldots, v_q}^{m}) \cdot \overbrace{\frac{y_{v_p}}{\kappa} \cdots \frac{y_{v_q}}{\kappa}}^{m}
\right)\!\!,\!
\label{eq:weighting:function}
\end{equation}
where $p, q$ denote vertex indices.

It is intuitive to enforce that ${\cal C}$ must contain at least one hyperedge, thus $\kappa$ must $\geq m$. Let $\delta_p = \frac{y_p}{\kappa}$ and $\epsilon = \frac{1}{\kappa}$. The conditions $\sum_{p=1}^{|V|} y_p = \kappa$ is then $\sum_{p=1}^{|V|} \delta_p = 1$. We relax the constraint of $y_p \in \{ 0, 1 \}$ to be $\delta_p \in [0, \epsilon]$, so $\delta$ is a continuous variable for optimization.
Eq.\eqref{eq:orig:hyper:objective} is re-written as:
\begin{equation}
\label{eq:hyper:objective}
\begin{split}
&\max_{x} \sum_{v_p,\ldots,v_q \in V}
\left(
W(  \overbrace{v_p, \ldots, v_q}^{m} ) \cdot \overbrace{ \delta_{v_p} \cdots \delta_{v_q} }^{m}
\right)
\\
& \;\;\;\;\;\; \textrm{s.t.}~\sum_{p=1}^{n} \delta_p = 1, \delta_p \in [0,\epsilon], \delta_r = \epsilon.
\end{split}
\end{equation}
We solve Eq.\eqref{eq:hyper:objective} using the pairwise update algorithm in \cite{Robust:Clustering}.

\subsection{Activity-Aware Tracking}
\label{sec:stage1}

{\bf Stage 1} of our method simultaneously recognizes individual activities and links tracklets in the following two steps (see Fig.\ref{fig:tracking:scheme} for a schematic overview):



\begin{itemize} \itemsep0em
\item \textbf{(T1) Generate candidate tracklets $\bar{X}$} from new detections $D$ that maximizes $\log f_{1}(X,D)$ in Eq.\eqref{eq:stage1:objective}.
\item \textbf{(T2) Link tracklets $X^{\prime}$ with $\bar{X}$} by maximizing the appearance, motion, and geometric consistencies that maximizes $\log \varphi_1(X,A)$ in Eq.\eqref{eq:stage1:objective}.
\end{itemize}

\noindent {\bf (T1) Generate candidate tracklets $\bar{X}$.}
For each existing target $\texttt{x}^{\prime}_{i} \in X^{\prime}$, we generate a set of candidate tracklets ${\bf \bar{x}}_{i} = \{ \bar{\texttt{x}}_{i,1}, \ldots, \bar{\texttt{x}}_{i,n} \}$ from observed detections $D$ using the tracking method in \cite{H2T}.~\footnote{
All candidate tracklets and their labels are denoted with a {\em bar} $\bar{\cdot}$.
}
We employ a gating strategy to restrict the number of candidate tracklets to consider. The appearance similarity between $\texttt{x}^{\prime}_{i}$ and each tracklet $\bar{\texttt{x}}_{j} \in \bar{X}$ is calculated using the POI features \cite{POI} and Euclidean metric. If this similarity is above a threshold $\theta_a$, $\bar{\texttt{x}}_{j}$ is added into ${\bf \bar{x}}_{i}$. Targets with no associated detection within time $[t-\tau - \tau_{a}, t-\tau-1]$ are discarded to reduce unnecessary computation. We use $\theta_a = 0.025$ and $\tau_{a} = 5~sec$ to include a rich set of candidate tracklets for linking. If any tracklet in $\bar{X}$ ends up not linked with any target (\eg, $\bar{\texttt{x}}_{5}$ in Fig.\ref{fig:tracking:scheme}), a new target is created. If any target $\texttt{x}^{\prime}_{i}$ ends up with no linked tracklet for status update, it is considered occluded.~\footnote{
We use trajectory prediction based on motion extrapolation in step (\textbf{R1}) of $\S$~\ref{sec:stage2} to determine if the target is still within the scene.
}

\noindent {\bf (T2) Link tracklets $X^{\prime}$ with $\bar{X}$.}
After candidate tracklets $\bar{X}$ are generated, for each candidate tracklet $\bar{\texttt{x}}_{i} \in \bar{X}$, we determine its individual activity label $\bar{\texttt{a}}_{i} \in \bar{A}$ for the purpose of activity-aware tracking. We consider $n_A$=$3$ individual activity labels regarding the motion pattern: \emph{standing}, \emph{walking}, and \emph{running}, by calculating the velocity $\bar{\nu}_{i}$ of each $\bar{\texttt{x}}_{i}$ and modeling the posteriors using sigmoid similar to \cite{ChangICCV11}: $p(\bar{\texttt{a}}_{i}|\bar{\nu}_{i}) \simeq p(\bar{\nu}_{i}|\bar{\texttt{a}}_{i})p(\bar{\texttt{a}}_{i})$. We consider social contextual cues and the correlations between individual activities in finding the best tracklet linking combinations. This also enables robust occlusion recovery for tracking.
Our solution is to represent all terms using a {\bf tracking hypergraph} $\mathcal{H}_\mathcal{T}$.
The clique potential function $\varphi_1(X,A)$ in Eq.\eqref{eq:stage1:objective} can then be inferred as
$\varphi_1(X,A) \sim \sum_{\forall \mathcal{C}^{\mathcal{T}}_{\alpha}} \Psi(\mathcal{C}^{\mathcal{T}}_{\alpha})$,
where $\mathcal{C}^{\mathcal{T}}_{\alpha}$ represents a cohesive cluster obtained from $\mathcal{H}_\mathcal{T}$, and $\alpha$ denotes cluster index.

The activity-aware tracking by linking tracklets $X'$ with $\bar{X}$ is performed in three sub-steps:
(\textbf{T2.1}) {\bf Estimate social group structure} using correlations between individual activities in a graph representation.
(\textbf{T2.2}) {\bf Construct hypergraph ${\cal H}_\mathcal{T}$}.
(\textbf{T2.3}) {\bf Optimize tracking} based on $\mathcal{H}_\mathcal{T}$.


\begin{figure*}[t]
\centerline{
  \includegraphics[width=1\linewidth]{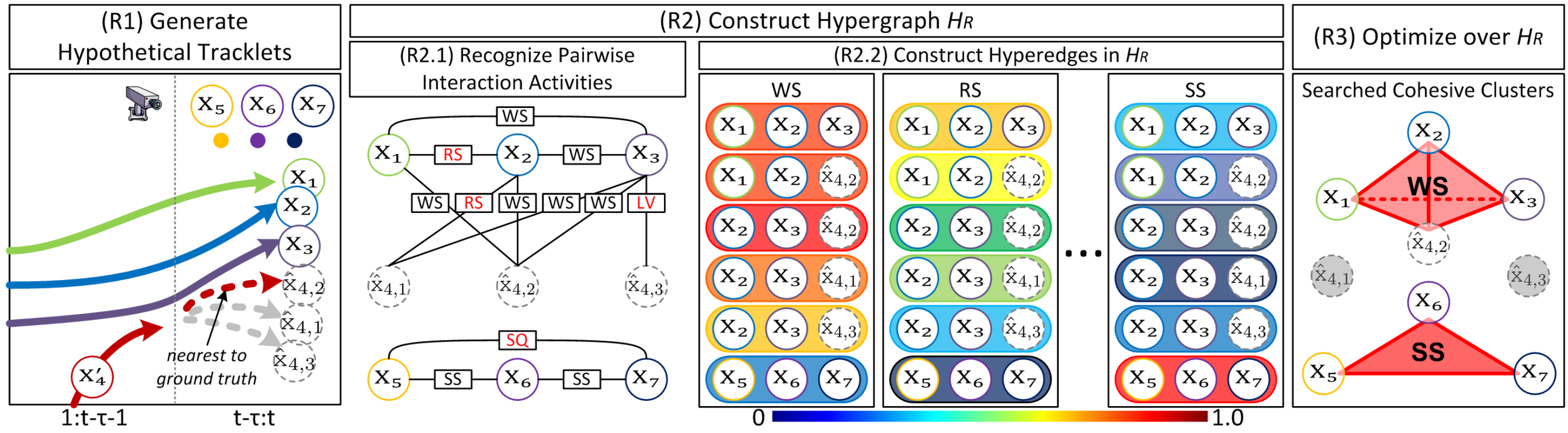} 
  \vspace{-0.1cm}
}
\caption{\em
{\bf Stage 2 joint interaction recognition and occlusion recovery} in a road-crossing scenario, where $\texttt{x}_{1}$, $\texttt{x}_{2}$, $\texttt{x}_{3}$ and $\texttt{x}_{4}^{\prime}$ are walking side-by-side across a road, while $\texttt{x}_{5}$, $\texttt{x}_{6}$, $\texttt{x}_{7}$ are standing side-by-side waiting.
Step {\bf (R1)} considers the linking of the occluded target $\texttt{x}^{\prime}_{4}$ to three hypothetical tracklets $\hat{\texttt{x}}_{4,1}, \hat{\texttt{x}}_{4,2}, \hat{\texttt{x}}_{4,3}$.
Step {\bf (R2)} constructs hypergraph $\mathcal{H}_\mathcal{R}$ for the inference in two steps.
(R2.1) evaluates each pairwise interaction by calculating a confidence score, where wrongly assigned labels are depicted in red.
(R2.2) constructs hyperedges based on the recognized pairwise interactions, where each hyperedge characterizes the likelihood of a pairwise interaction.
Step {\bf (R3)} optimizes the inference over ${\cal H}_\mathcal{R}$ to jointly recognize interaction labels and resolve the tracklet linking and occlusion recovery.
\vspace{-0.4cm}
}
\label{fig:recog:scheme}
\end{figure*}

\begin{figure}[t]
\centerline{
  \includegraphics[width=0.6\linewidth]{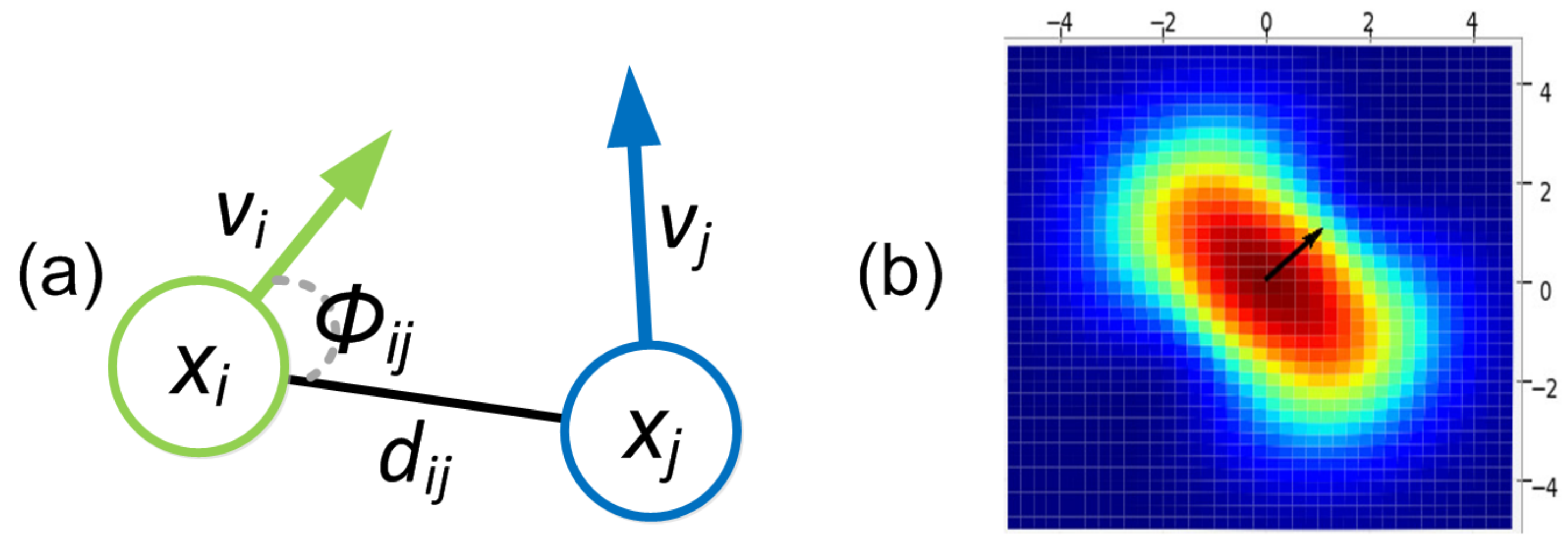}\\ 
}
\caption{\em
Social group affinity between a pair of individuals is calculated based on: (a) distance, angle, and motion (velocity magnitude \& direction).
(b) visualizes such a measure at (0, 0) with direction vector (1, 1) arrow in a color map depicting the probability kernel between 0 (blue) and 1 (red).
}
\label{fig:group_kernel}
\end{figure}

{\bf (T2.1) Estimate social group structure.}
We represent the social group structure of tracked targets and the correlations between individual activities using an {\em undirected complete graph} $\mathcal{\tilde{G}} = \{ \tilde{V}, \tilde{E}, \tilde{W} \}$ with $\tilde{V} = X^{\prime}$. $\forall \texttt{x}_{i}^{\prime}, \texttt{x}_{j}^{\prime}, \exists e_{ij} = (\texttt{x}_i^{\prime}, \texttt{x}_j^{\prime}) \in \tilde{E}$. Edge weight $\tilde{W}(e_{ij})$ reflects the correlation between activities $\texttt{a}_i$ and $\texttt{a}_j$ of $\texttt{x}_i^{\prime}$ and $\texttt{x}_j^{\prime}$, respectively. We define $p_{corr}$ to reflect the correlation between activities of two targets similar to \cite{ChangICCV11}:
\begin{equation}
\label{eq:individual:correlation:measure}
p_{corr}(\texttt{x}_{i}, \texttt{x}_{j}) = g(d_{ij}, \phi_{ij}, \| \nu_{i} \|, \| \nu_{j} \|, \texttt{a}_i, \texttt{a}_j, \eta(\texttt{x}_{i}), \eta(\texttt{x}_{j})),
\end{equation}
where $d_{ij}$ represents Euclidean distance between the targets.
As shown in Fig.\ref{fig:group_kernel}a, $\phi_{ij}$ represents the angle between the facing direction of $\texttt{x}_{i}$ and the relative vector from $\texttt{x}_{i}$ to $\texttt{x}_{j}$, and $\nu_{i}$ represents the velocity of $\texttt{x}_{i}$.
For a target $\texttt{x}_{i}$, if $\texttt{a}_{i}$ is recognized as ``standing", we use the classifier in \cite{CAD:umich} to calculate the body orientation $\eta(\texttt{x}_i)$ out of 8 quantizations. Otherwise, $\eta(\texttt{x}_{i})$ estimates motion direction from the trajectory.
Edge weights of $\mathcal{\tilde{G}}$ are calculated according to Eq.\eqref{eq:individual:correlation:measure} and refined using further grouping cues as in \cite{ChangICCV11}. Fig.\ref{fig:group_kernel}b visualizes the correlation defined by Eq.\eqref{eq:individual:correlation:measure}. The probability is higher on the side of a person than in the front or back, which is an implementation of Hall's {\em proxemics} social norms \cite{The:hidden:dimension}. We discard edges with weights lower than $0.3$ to obtain a sparse graph denoted as $\mathcal{\tilde{G}}_s$ for computation speedup.

{\bf (T2.2) Construct hypergraph ${\cal H}_\mathcal{T} = \{ V_{\mathcal{T}}, E_{\mathcal{T}}, W_{\mathcal{T}} \}$ using $\mathcal{\tilde{G}}_s$} to capture the high-order correlations between activities within a group.
A vertex $v_p \in V_{\mathcal{T}}$ represents a hypothesis of linking a tracked target with its candidate tracklet, \ie, $v_p = \texttt{x}^{\prime}_{i} \oplus \bar{\texttt{x}}_{i,k}$ where ``$\oplus$" represents the association of two tracklets.
A $m$-degree hyperedge $\mathbf{e}^{m} \in E_{\mathcal{T}}$ represents the combination of $m$ tracklet linking hypotheses in an assignment.

The linking of tracklets $X'$ with $\bar{X}$ can be considered as an assignment problem with the following two {\bf tracklet assignment constraints}: {\em (i)} a target cannot be linked with two or more candidate tracklets, and {\em (ii)} a candidate tracklet cannot be linked with two or more targets.
We enforce these constraints in the construction of hyperedges in $\mathcal{H}_\mathcal{T}$.
Specifically, $\forall v_p, v_q \in V_{\mathcal{T}}$, where $v_p = \texttt{x}^{\prime}_{i} \oplus \bar{\texttt{x}}_{i,k}$ and $v_{q} = \texttt{x}^{\prime}_{j} \oplus \bar{\texttt{x}}_{j,l}$, \emph{if and only if} $e_{ij} = (\texttt{x}^{\prime}_{i}, \texttt{x}^{\prime}_{j}) \in {\cal \tilde{G}}_s$, $v_p$ and $v_q$ can co-exist in a hyperedge in $\mathcal{H}_\mathcal{T}$.

We further consider motion and behavior consistencies and their correlations (via ${\cal \tilde{G}}_s$) in determining the hyperedge weights. Specifically, we consider three affinities that determine the hyperedge weights: the {\em appearance} ($W_a$) of each tracklet, the {\em facing-direction} ($W_d$) and the {\em geometric similarity} ($W_g$) between tracked targets.

The appearance affinity between a target $\texttt{x}_i^{\prime}$ and a candidate tracklet $\bar{\texttt{x}}_{i,k}$ is computed using the appearance features of tracklets as \cite{POI}:
\begin{equation}
\label{eq:app:affinity:h1}
W_{a}(\mathbf{e}^{m}) = \sum_{\texttt{x}^{\prime}_{i} \oplus \bar{\texttt{x}}_{i,k} \in \mathbf{e}^{m}} |\texttt{x}^{\prime}_{i} - \bar{\texttt{x}}_{i,k}|.
\end{equation}

We assume that activity states (such as walking direction) do not change abruptly in-between small linked tracklets. In other words, difference between facing directions of two targets should be small for linked tracklets:
\begin{equation}
\label{eq:dr:affinity:h1}
W_{d}(\mathbf{e}^{m}) = \sum_{\texttt{x}^{\prime}_{i} \oplus \bar{\texttt{x}}_{i,k} \in \mathbf{e}^{m}} \cos(\eta(\texttt{x}^{\prime}_{i}),\eta(\bar{\texttt{x}}_{i,k})).
\end{equation}

Our method aims to run on surveliiance videos without calibration. To ensure smooth tracking, we use a geometric affinity term $W_{g}$ to ensure that relative angles between two targets does not change abruptly:
\begin{equation}
\label{eq:geom:affinity:h1}
W_{g}(\mathbf{e}^{m}) = \sum_{\texttt{x}^{\prime}_{i} \oplus \bar{\texttt{x}}_{i,k} \in \mathbf{e}^{m}} \sum_{\texttt{x}^{\prime}_{j} \oplus \bar{\texttt{x}}_{j,l} \in \mathbf{e}^{m}} \cos(\mathbf{p}^{\prime}_{ij}, \mathbf{\bar{p}}_{ij}),
\end{equation}
where $\mathbf{p}^{\prime}_{ij}$ and $\mathbf{\bar{p}}_{ij}$ represent the relative image coordinate vectors between tracked targets and candidate tracklets.
Final affinity value of a hyperedge $\mathbf{e}^{m}$ is computed by
$W(\mathbf{e}^{m}) = \lambda_a W_{a}(\mathbf{e}^{m}) + \lambda_d W_{d}(\mathbf{e}^{m}) + \lambda_g W_{g}(\mathbf{e}^{m})$,
where $\lambda_a$, $\lambda_d$, $\lambda_g$ are set as $\lambda_a = 30$, $\lambda_d = 1$, $\lambda_g = 0.5$.

{\bf (T2.3) Optimize tracking} based on ${\cal H}_\mathcal{T}$.
This step aims to determine the optimal tracklet linking among candidates represented in the hypergraph $\mathcal{H}_\mathcal{T}$. The optimization is performed by the cohesive cluster search on $\mathcal{H}_\mathcal{T}$ described in $\S$~\ref{sec:hypergraph:opt}.
For each vertex $v_r$, such a search yields a cluster ${\cal C}(v_r)$ with a score. Since a vertex may appear in multiple clusters, if any resulting cluster violates the {\em tracklet assignment constraints} in ({\bf T2.2}), such a cluster is removed to avoid further consideration. We ensure that the resulting cohesive clusters represent valid tracklet linking hypotheses that is sound and redundancy-free. \footnote{
Hypergraph clusters are processed sequentially in descending order of their scores. If any cluster violates the constraints,
new cluster is discarded and any duplication is removed.
}
In case a target ends up not linked with any candidate tracklets (\eg, $\texttt{x}^{\prime}_{5}$ in Fig.\ref{fig:recog:scheme}), such a target should be either outside the scene or under occlusion. We store all discovered occlusions and will try to recover them at {\bf Stage 2} in $\S$ \ref{sec:stage2}.
Finally, target trajectories $X$ are updated with the newly linked tracklets in $\bar{X}$ to be $X^*$, and activity labels $A$ are augmented with respective ones in $\bar{A}$ to be $A^*$.


\subsection{Joint Interaction Recognition and Occlusion Recovery}
\label{sec:stage2}

Our approach is motivated from the observation that pairwise interactions $I$ within a group can provide rich contextual cues to recognize the activities (as in Fig.\ref{fig:overview}) and recover possible occlusions. \textbf{Stage~2} of our method jointly resolves the two problems of (1) recognizing pairwise interactions $I$ and (2) occlusion recovery to improve tracking. We again use a hypergraph representation to explore the high-order correlations among the interactions $I$, such that a similar cluster search scheme can be applied for optimization. Specifically, we construct the (activity) {\bf recognition hypergraph} ($\mathcal{H}_\mathcal{R}$) based on the inferred target locations $X^*$ and individual activities $A^*$.
The optimization over $\mathcal{H}_\mathcal{R}$ maximizes the clique potential function $\log \varphi_2(A^*, I) \varphi_3(I, X^*)$ in Eq.\eqref{eq:stage2:objective}, as $\varphi_2(A^*, I)\varphi_3(I, X^*) \sim \sum_{\forall \mathcal{C}_{\alpha}^{\mathcal{R}}} \Psi(\mathcal{C}_{\alpha}^{\mathcal{R}})$,
where $\mathcal{C}_{\alpha}^{\mathcal{R}}$ represents a cohesive cluster obtained from $\mathcal{H}_\mathcal{R}$.

\textbf{Stage~2} of our method jointly recognizes $I$ and recovery occlusions in the following three main steps (see Fig.\ref{fig:recog:scheme} for a schematic overview):
\begin{itemize} \itemsep0em

\item \textbf{(R1) Generate hypothetical tracklets $\hat{X}$} for occlusion recovery from given existing $X^{\prime}$ and $A^{\prime}$.

\item \textbf{(R2) Construct hypergraph} ${\cal H}_\mathcal{R}$ based on $X^*, A^*, \hat{X}$ to infer high-order correlations among their pairwise interactions $I$.

\item \textbf{(R3) Optimize recognition and recovery over ${\cal H}_\mathcal{R}$} to simultaneously recognize interaction $I$ and link occluded targets with suitable hypothetical tracklets.

\end{itemize}

\noindent {\bf (R1) Generate hypothetical tracklets $\hat{X}$.}
For each possibly occluded target $\texttt{x}^{\prime}_{i} \in X^{\prime}$, we generate a few hypothetical tracklets $\mathbf{\hat{x}}_{i} = \{ \hat{\texttt{x}}_{i,1}, \ldots, \hat{\texttt{x}}_{i,h} \}$ based on trajectory predictions, where $h$ is empirically set to $9$.~\footnote{
All hypothetical tracklets are denoted with a {\em hat} $\hat{\cdot}$ across the paper.
}
For a moving target $\texttt{x}_{i}^{\prime}$ with $\texttt{a}_{i}^{\prime} = walking$, we generate $\mathbf{\hat{x}}_{i}$ via motion extrapolation. For a stationary target $\texttt{x}_{i}^{\prime}$ with $\texttt{a}_{i}^{\prime} = standing$, we add a small perturbation to $\mathbf{\hat{x}}_{i}$.

\noindent {\bf (R2) Construct hypergraph} $\mathcal{H}_\mathcal{R} = \{V_\mathcal{R}, E_\mathcal{R}, W_\mathcal{R}\}$, such that high-order correlations among interactions among $X$ and $\hat{X}$ are captured for the purposes of simultaneous activity recognition and occlusion recovery. Thus, $V_\mathcal{R} = X \cup \hat{X}$.
Each hyperedge in $E_\mathcal{R}$ characterizes the likelihood of a pairwise interaction $\texttt{i} \in I$. For example in Fig.\ref{fig:recog:scheme}, $\texttt{x}_{1}, \texttt{x}_{2}, \texttt{x}_{3}$ are connected by 3 hyperedges, which correspond to interactions ``WS", ``RS", ``SS", respectively. See  $\S$ \ref{sec:experiments} for a complete list of interaction class defined in public datasets \cite{CAD:umich,Choi:Savarese:Tracking:Collective:Activity:Reog:ECCV2012}. We denote $n_I$ the number of interaction classes.

The inference of each interaction class can be optimized independently. We can thus decompose $\mathcal{H}_\mathcal{R}$ into $n_I$ sub-hypergraphs $\{\mathcal{\breve{H}}_{\beta}\}_{\beta=1}^{n_I}$, with $\mathcal{\breve{H}}_{\beta} = \{V_\mathcal{R}, \breve{E}_{\beta}, \breve{W}_{\beta}\}$ for the $\beta$-th interaction class. For each hyperedge $\mathbf{e}^{m} \in \breve{E}_{\beta}$, the weight $\breve{W}_{\beta}(\mathbf{e}^{m})$ reflects how likely the interaction between the $m$ targets are cohesive as a whole (\eg, all walking-side-by-wide).

We calculate the hyperedge weights in ${\cal H}_\mathcal{R}$ in two steps: {\bf (R2.1)} evaluates each pairwise interaction with a confidence score. {\bf (R2.2)} constructs hyperedges in ${\cal H}_\mathcal{R}$ using the average score from all involved targets.

{\bf (R2.1) Recognize pairwise interaction activities.}
We calculate a confidence score for each possible pairwise interaction $\texttt{i}_{ij}$ between the targets $\texttt{x}_{i}$, $\texttt{x}_{j}$ using a simple effective rule-based probabilistic approach as in \cite{ChangICCV11}. Specifically, the confidence score of $\texttt{i}_{ij}$ belonging to the $\beta$-th class is calculated by multiplying the following six component probabilities: {\em distance} (ds), {\em group connectivity} (gc) calculated in~\eqref{eq:individual:correlation:measure}, {\em individual activity agreement} (aa), {\em distance change type} (dc), {\em facing direction} (dr), and {\em frontness/sidedness} (fs):
\begin{equation}
\begin{split}
&\hspace{-0.1cm}p(\texttt{i}_{ij} = \beta | \texttt{x}_{i},\texttt{x}_{j},\texttt{a}_{i},\texttt{a}_{j}) =
  p_{ds}(\beta | \texttt{x}_{i}, \texttt{x}_{j}) \cdot p_{gc}( \beta | \texttt{a}_{i}, \texttt{a}_{j}) \cdot \\
& \;\; p_{aa}( \beta | \texttt{a}_{i}, \texttt{a}_{j}) \cdot p_{dc}( \beta | \texttt{x}_{i}, \texttt{x}_{j}) \cdot p_{dr}( \beta | \texttt{x}_{i}, \texttt{x}_{j}) \cdot p_{fs}( \beta | \texttt{x}_{i}, \texttt{x}_{j}).
\end{split}
\label{eq:interaction:prob}
\end{equation}
Detailed formulation of the above component probabilities and formulation are provided in Tables~\ref{tab:component:probs} and \ref{tab:interaction:formulas}.


\begin{table*}[t]
\caption{\em Component probabilities for the pairwise interaction activities. The parameters used in these component probabilities, \eg, the means and standard deviations are calculated from the training dataset.
\vspace{-0.3cm}
}
\label{tab:component:probs}
\centering
\begin{tabular}{|p{12em}||p{47.5em}|}
\hline
  \makebox[12em]{Component} & \makebox[47.5em]{Probability}\\ \hline \hline
  \makebox[12em]{Distance} & $p_{ds}( \emph{within-effective-range} |x_{i},x_{j}) = \delta(|\frac{d_{ij}-\mu_{ds}}{\sigma_{ds}}| \leq b ), d_{ij} \sim \mathcal{N}(\mu_{ds}, \sigma_{ds})$, where $\mathcal{N}$ denotes normal distribution \\
\hline
  \multirow{3}{*}{\makebox[12em]{Group connectivity}}
  & $p_{gc}( \texttt{GC} |x_{i},x_{j})$, where $\texttt{GC} \in \{ \emph{connect}, \emph{not-connect} \}$\\
  & $p_{gc}( \emph{connect} |x_{i},x_{j}) = g(d_{ij}, \phi_{ij}, \| \nu_{i} \|, \| \nu_{j} \|)$, where $g(\cdot)$ is defined in Eq.\eqref{eq:individual:correlation:measure}\\
  & $p_{gc}( \emph{not-connect} |x_{i},x_{j}) = 1-g(d_{ij}, \phi_{ij}, \| \nu_{i} \|, \| \nu_{j} \|)$ \\
\hline
  \makebox[12em]{Individual activity agreement} & $p_{aa}(a_{i} = \texttt{AA}_{1}, a_{j} = \texttt{AA}_{2}) = \sqrt{p(\texttt{AA}_{1}|\nu_{i}) \cdot p(\texttt{AA}_{2}|\nu_{j})}$, where $\texttt{AA}_{1}, \texttt{AA}_{2} \in \{\emph{standing}, \emph{walking}, \emph{running}\}$ \\
\hline
  \multirow{5}{*}{\makebox[12em]{Distance-change type}}
  & $p_{dc}(\texttt{DC} | x_{i}, x_{j})$, where $\texttt{DC} \in \{ \emph{decreasing}, \emph{unchanging}, \emph{increasing} \}$ \\
  & $p_{dc}(\emph{decreasing} | x_{i}, x_{j}) = 1 - sigmoid(d^{g}_{ij}, \mu_{d2u}, \sigma_{d2u})$, where $d^{g}_{ij} = d_{ij}-d^{\prime}_{ij}$ \\
  & $p_{dc}(\emph{unchanging} | x_{i}, x_{j}) = \lambda * sigmoid(d^{g}_{ij}, \mu_{d2u}, \sigma_{d2u}) + (1-\lambda) * ( 1-sigmoid(d^{g}_{ij}, \mu_{u2i}, \sigma_{u2i}) )$; \\
  & if $d^{g}_{ij} < \mu_{d2u}$, $\lambda = 1$; if $d^{g}_{ij} \geq \mu_{u2i}$, $\lambda = 0$; otherwise, $\lambda = 1-\frac{ d^{g}_{ij}- \mu_{d2u} }{ \mu_{u2i} - \mu_{d2u} }$\\
  & $p_{dc}(\emph{increasing} | x_{i}, x_{j}) = sigmoid(d^{g}_{ij}, \mu_{u2i}, \sigma_{u2i})$ \\
\hline
  \makebox[12em]{Facing direction} & $p_{dr}( \texttt{DR} | x_{i}, x_{j}) \in \{0,1\}$, where $\texttt{DR} \in \{ \emph{same}, \emph{opposite}, \emph{frequent-changing} \}$ \\ \hline
  \multirow{3}{*}{\makebox[12em]{Frontness/sideness}}
  & $p_{fs}( \texttt{FS} | x_{i}, x_{j} )$, where $\texttt{FS} \in \{ \emph{frontness}, \emph{sideness} \}$  \\
  & $p_{fs}( \emph{frontness} | x_{i}, x_{j} ) = \max(cos(\mathbf{p}_{ij}, x_{i}), cos(\mathbf{p}_{ij}, x_{j}))$\\
  & $p_{fs}( \emph{sideness} | x_{i}, x_{j} ) = 1-p_{fs}( \emph{frontness} | x_{i}, x_{j} )$\\
\hline
\end{tabular}
\vspace{-0.2cm}
\end{table*}

\begin{table*}[t]
\caption{\em Probabilistic formulations for the pairwise interactions $p(\texttt{i}_{ij}=\beta)$. We define {\em dancing-together} (DT) as a new interaction activity class to deal with the new collective activity ``dancing'' in the Augmented-CAD.
\vspace{-0.3cm}
}
\label{tab:interaction:formulas}
\centering
\begin{tabular}{|c|c|c|} 
\hline
  Pairwise Interaction $p(\texttt{i}_{ij}=\beta)$  & Associated Collective Activity (C) & Probabilistic Formulation \\
\hline \hline
  {\em facing-each-other} &
  \multirow{2}{*}{\sc talking}
  & $p_{ds}(\emph{within-effective-range}) \cdot p_{gc}(\emph{connect}) \cdot p_{aa}(\emph{standing},\emph{standing}) \cdot $\\
  ($\beta = $FE) & & $p_{dc}( \emph{unchanging} ) \cdot p_{dr}( \emph{opposite} ) \cdot p_{fs}( \emph{frontness} )$  \\
\hline
  {\em standing-in-a-row} &
  \multirow{2}{*}{\sc queuing}
  & $p_{ds}(\emph{within-effective-range}) \cdot p_{gc}(\emph{connect}) \cdot p_{aa}(\emph{standing},\emph{standing}) \cdot $\\
  ($\beta = $SR) & & $p_{dc}( \emph{unchanging} ) \cdot p_{dr}( \emph{same} ) \cdot p_{fs}( \emph{frontness} )$  \\
\hline
  {\em standing-side-by-side} &
  \multirow{2}{*}{\sc waiting}
  & $p_{ds}(\emph{within-effective-range}) \cdot p_{gc}(\emph{connect}) \cdot p_{aa}(\emph{standing},\emph{standing}) \cdot $\\
  ($\beta = $SS) & & $p_{dc}( \emph{unchanging} ) \cdot p_{dr}( \emph{same} ) \cdot p_{fs}( \emph{sideness} )$  \\
\hline
  {\em dancing-together} &
  \multirow{2}{*}{\sc dancing}
  & $p_{ds}(\emph{within-effective-range}) \cdot p_{gc}(\emph{connect}) \cdot p_{aa}(\emph{walking},\emph{walking}) \cdot $\\
  ($\beta = $DT) & & $p_{dc}( \emph{unchanging} ) \cdot p_{dr}( \emph{frequent-chaning} ) \cdot p_{fs}( \emph{sideness} )$  \\
\hline
  {\em approaching} &
  \multirow{2}{*}{\sc gathering}
  & $p_{ds}(\emph{within-effective-range}) \cdot p_{gc}(\emph{not-connect}) \cdot p_{aa}(\emph{walking},\emph{walking}) \cdot $\\
  ($\beta = $AP) & & $p_{dc}( \emph{decreasing} ) \cdot p_{dr}( \emph{opposite} ) \cdot p_{fs}( \emph{frontness} )$  \\
\hline
  {\em walking-in-opposite-} &
  \multirow{3}{*}{\sc dismissal}
  & $p_{ds}(\emph{within-effective-range}) \cdot p_{gc}(\emph{not-connect}) \cdot p_{aa}(\emph{walking},\emph{walking}) \cdot $\\
  directions ($\beta = $WO), & & $p_{dc}( \emph{increasing} ) \cdot p_{dr}( \emph{opposite} ) \cdot p_{fs}( \emph{frontness} )$  \\
  {\em leaving} ($\beta = $LV) & & \\
\hline
  {\em walking-side-by-side} &
  {\sc crossing}
  & $p_{ds}(\emph{within-effective-range}) \cdot p_{gc}(\emph{connect}) \cdot p_{aa}(\emph{walking},\emph{walking}) \cdot $\\
  ($\beta = $WS) & {\sc walking (together)} & $p_{dc}( \emph{unchanging} ) \cdot p_{dr}( \emph{same} ) \cdot p_{fs}( \emph{sideness} )$  \\
\hline
  {\em running-side-by-side} &
  \multirow{2}{*}{\sc jogging}
  & $p_{ds}(\emph{within-effective-range}) \cdot p_{gc}(\emph{connect}) \cdot p_{aa}(\emph{running},\emph{running}) \cdot $\\
  ($\beta = $RS) & & $p_{dc}( \emph{unchanging} ) \cdot p_{dr}( \emph{same} ) \cdot p_{fs}( \emph{sideness} )$  \\
\hline
  {\em running-one-after-the-other} &
  \multirow{2}{*}{\sc chasing}
  & $p_{ds}(\emph{within-effective-range}) \cdot p_{gc}(\emph{connect}) \cdot p_{aa}(\emph{running},\emph{running}) \cdot $\\
  ($\beta = $RR) & & $p_{dc}( \emph{unchanging} ) \cdot p_{dr}( \emph{same} ) \cdot p_{fs}( \emph{frontness} )$  \\
\hline
\end{tabular}
\vspace{-0.2cm}
\end{table*}

{\bf (R2.2) Construct hyperedges in ${\cal H}_\mathcal{R}$.}
We consider interactions among both real and hypothetical targets during the optimization.
We avoid the inclusion of multiple hypothetical tracklets of a target into a hyperedge.
For each hyperedge $\mathbf{e}^{m} = \{ \texttt{x}_1^{\mathbf{e}}, \ldots, \texttt{x}_m^{\mathbf{e}} \} \in E_{\beta}$ for the $\beta$-th interaction class, we calculate the edge weight by averaging the confidence scores of the involved targets:
\begin{equation}
\label{eq14}
W(\mathbf{e}^{m}) = \frac{1}{\binom{m}{2}} \sum_{i,j} p(\texttt{i}_{ij} = \beta | \texttt{x}^{\mathbf{e}}_{i},\texttt{x}^{\mathbf{e}}_{j},\texttt{a}^{\mathbf{e}}_{i},\texttt{a}^{\mathbf{e}}_{j}).
\end{equation}

\noindent {\bf (R3) Optimize recognition and recovery over ${\cal H}_\mathcal{R}$}
cohesive cluster search on each sub-hypergraph $\mathcal{\breve{H}}_{\beta}$ respectively (as described in $\S$~\ref{sec:hypergraph:opt}).
This optimizes the assignment of interaction labels and the linking of probable hypothetical tracklets.
Similar to (\textbf{T2.3}), for each vertex $v_r \in {\cal H}_\mathcal{R}$, we search for candidate cohesive clusters with confidence scores.
We ensure that the resulting cohesive clusters are sound and redundancy-free, also not violating the {\em tracklet assignment constraints}.
%
Optimization results are used to update $X^*, A^*, I$ into $X^{\ddagger}, A^{\ddagger}, I^*$, respectively as in Eq.\eqref{eq:stage2:objective}.

\subsection{Collective Activity Recognition}
\label{sec:stage3}

\textbf{Stage~3} of our method infers the collective activities $C^*$ for each individual in a group, based on an intuition that collective activity is characterized by pairwise interactions $I$ indexed by $\beta$ within the group. Fig.\ref{fig:overview} illustrates several examples, and Table~\ref{tab:interaction:formulas} shows the cohesive pairwise interaction for each collective activity class. For each target $\texttt{x}_{i}$ within a group, we infer the most probable collective activity. The term $\log f_3(X^{\ddagger},A^{\ddagger},I^*,C)$ in Eq.\eqref{eq:stage3:objective} can be maximized based on a probabilistic formulation similar to \cite{ChangICCV11}. Consider the $\beta$-th interaction for the $c$-th collective activity, $p (c|\texttt{x}_{i})$ is calculated using:
\vspace{-0.2cm}
\begin{equation}
\label{eq:collective:prob}
p (c|\texttt{x}_{i}) = 1 - \prod_{\forall j} \left( 1 - p(\texttt{i}_{ij} = \beta | \texttt{x}_{i},\texttt{x}_{j}) \right),
\vspace{-0.1cm}
\end{equation}
where $p(\texttt{i}_{ij} = \beta | \texttt{x}_{i},\texttt{x}_{j})$ is obtained in Eq.(13) after the optimization in (\textbf{R3}).
The collective activity of $\texttt{x}_{i}$ is determined by $\argmax_{c} \; p(c|\texttt{x}_{i})$.
We use the collective activity involving most participants as the label of the scene, to comply with the practice in major datasets \cite{Choi:Savarese:Tracking:Collective:Activity:Reog:ECCV2012,CAD:umich,Augmented:CAD:umich}.~\footnote{
If there are insufficient targets for interactive or collective activities ({\em e.g.} people leaving the scene), we keep existing labels for a short while.
}

\section{Experiments}
\label{sec:experiments}

\begin{table*}[t]
\caption{\em \small Activity recognition evaluation results in terms of accuracy and comparison with state-of-the-art methods. The CAD results are splitted into two tables: one for comparing methods only producing collective activity results, and the other for all three activities.
\vspace{-0.3cm}
}
\centerline{
    \begin{minipage}[t]{0.5\hsize}\centering
        \begin{tabular}[t]{|p{8.2em}|p{1.5em}|p{1.5em}|p{1.5em}|p{1.5em}|p{1.5em}|p{1.5em}|} \hline
          \multicolumn{7}{|c|}{CAD \cite{CAD:umich}} \\ \hline
          \multirow{2}{*}{\makebox[5.4em]{Method}} & \multicolumn{2}{c|}{\makebox[4em]{Collective} }& \multicolumn{2}{c|}{\makebox[4em]{Interaction}} & \multicolumn{2}{c|}{\makebox[4em]{Individual}}\\ \cline{2-7}
          & \makebox[2em]{OCA} &\makebox[2em]{MCA} &\makebox[2em]{OCA} &\makebox[2em]{MCA} &\makebox[2em]{OCA} &\makebox[2em]{MCA}\\ \hline
          LCC \cite{Augmented:CAD:umich} & \makebox[1.5em]{-} & 70.9 & \makebox[1.5em]{-} & \makebox[1.5em]{-} & \makebox[1.5em]{-} & \makebox[1.5em]{-}\\ \hline
          CFT \cite{Combining:Per-frame:and:Per-track:Cues} & \makebox[1.5em]{-} & 75.1 & \makebox[1.5em]{-} & \makebox[1.5em]{-} & \makebox[1.5em]{-} & \makebox[1.5em]{-}\\ \hline
          FM \cite{Flow:Model} & \makebox[1.5em]{-} & 70.9 & \makebox[1.5em]{-} & \makebox[1.5em]{-} & \makebox[1.5em]{-} & \makebox[1.5em]{-}\\ \hline
          LOG \cite{Discriminative:Latent:Models} & 79.7 & 78.4 & \makebox[1.5em]{-} & \makebox[1.5em]{-} & \makebox[1.5em]{-} & \makebox[1.5em]{-}\\ \hline
          MCTS \cite{AndOrGraph:Scheduling:Activity:ICCV2013} & \makebox[1.5em]{-} & 88.9 & \makebox[1.5em]{-} & \makebox[1.5em]{-} & \makebox[1.5em]{-} & \makebox[1.5em]{-}\\ \hline
          LLC \cite{Learning:Latent:Constituents} & \makebox[1.5em]{-} & 75.1 & \makebox[1.5em]{-} & \makebox[1.5em]{-} & \makebox[1.5em]{-} & \makebox[1.5em]{-}\\ \hline
          HiRF \cite{HiRF:Group:Activity:ECCV2014} & \makebox[1.5em]{-} & 92.0 & \makebox[1.5em]{-} & \makebox[1.5em]{-} & \makebox[1.5em]{-} & \makebox[1.5em]{-}\\ \hline
          DSM \cite{Deep:Structured:Models} & \makebox[1.5em]{-} & 80.6 & \makebox[1.5em]{-} & \makebox[1.5em]{-} & \makebox[1.5em]{-} & \makebox[1.5em]{-}\\ \hline
          CK \cite{Visual:recognition:by:counting:instances} & 83.4 & 81.9 & \makebox[1.5em]{-} & \makebox[1.5em]{-} & \makebox[1.5em]{-} & \makebox[1.5em]{-}\\ \hline
          HDT \cite{Hierarchical:Deep:Temporal:Model} & \makebox[1.5em]{-} &  81.5 & \makebox[1.5em]{-} & \makebox[1.5em]{-} & \makebox[1.5em]{-} & \makebox[1.5em]{-}\\ \hline
          SIE \cite{Structure:Inference:Machines} & \makebox[1.5em]{-} & 81.2 & \makebox[1.5em]{-} & \makebox[1.5em]{-} & \makebox[1.5em]{-} & \makebox[1.5em]{-}\\ \hline
          RMI \cite{Recurrent:Modeling:Interaction:Context} & \makebox[1.5em]{-} & 89.4 & \makebox[1.5em]{-} & \makebox[1.5em]{-} & \makebox[1.5em]{-} & \makebox[1.5em]{-}\\ \hline
          UTR-1 \cite{Choi:Savarese:Tracking:Collective:Activity:Reog:ECCV2012,Choi:Collective:Activities:PAMI2014} & 79.0 & 79.6 & 56.2 & 50.8 & \makebox[1.5em]{-} & \makebox[1.5em]{-} \\ \hline
          UTR-2 \cite{Choi:Savarese:Tracking:Collective:Activity:Reog:ECCV2012,Choi:Collective:Activities:PAMI2014} & 79.4 & 80.2 & 45.5 & 36.6 & \makebox[1.5em]{-} & \makebox[1.5em]{-} \\ \hline
          Baseline & 87.8 & 88.0 &65.4  &48.4  &87.3  &88.3  \\ \hline
          Ours w/o ${\cal H}_{\mathcal{T}}$  &91.8  &91.8  &74.3  &55.6  &\textbf{87.7}  &88.4  \\ \hline
          Ours w/o ${\cal H}_{\mathcal{R}}$  & 88.0  & 87.8  &71.4  &56.5  &87.3  &88.4  \\ \hline
          Ours w/o ${\cal H}_{\mathcal{T}},{\cal H}_{\mathcal{R}}$  &87.9  &87.6  &67.3  &53.7  &87.4  &\textbf{88.6}  \\ \hline
          Ours & \textbf{92.5} & \textbf{92.4} &\textbf{78.1}  &\textbf{57.6}  &87.4  &88.1  \\ \hline
        \end{tabular}
    \end{minipage}
    \begin{minipage}[t]{0.5\hsize}\centering
    \begin{tabular}[t]{|p{8.2em}|p{1.5em}|p{1.5em}|p{1.5em}|p{1.5em}|p{1.5em}|p{1.5em}|} \hline
      \multicolumn{7}{|c|}{Augmented CAD \cite{Augmented:CAD:umich}} \\ \hline
      \multirow{2}{*}{\makebox[5.2em]{Method}} & \multicolumn{2}{c|}{\makebox[4em]{Collective} }& \multicolumn{2}{c|}{\makebox[4em]{Interaction}} & \multicolumn{2}{c|}{\makebox[4em]{Individual}}\\ \cline{2-7}
      & \makebox[2em]{OCA} &\makebox[2em]{MCA} &\makebox[2em]{OCA} &\makebox[2em]{MCA} &\makebox[2em]{OCA} &\makebox[2em]{MCA}\\ \hline
      LCC \cite{Augmented:CAD:umich} & \makebox[1.5em]{-} & 82.0 & \makebox[1.5em]{-} & \makebox[1.5em]{-} & \makebox[1.5em]{-} & \makebox[1.5em]{-} \\ \hline
      CFT \cite{Combining:Per-frame:and:Per-track:Cues} & \makebox[1.5em]{-} & 85.8 & \makebox[1.5em]{-} & \makebox[1.5em]{-} & \makebox[1.5em]{-} & \makebox[1.5em]{-} \\ \hline
      FM \cite{Flow:Model} & \makebox[1.5em]{-} & 83.7 & \makebox[1.5em]{-} & \makebox[1.5em]{-} & \makebox[1.5em]{-} & \makebox[1.5em]{-} \\ \hline
      LLC \cite{Learning:Latent:Constituents} & \makebox[1.5em]{-} & 90.1 & \makebox[1.5em]{-} & \makebox[1.5em]{-} & \makebox[1.5em]{-} & \makebox[1.5em]{-} \\ \hline
      SIE \cite{Structure:Inference:Machines} & \makebox[1.5em]{-} & 90.2 & \makebox[1.5em]{-} & \makebox[1.5em]{-} & \makebox[1.5em]{-} & \makebox[1.5em]{-} \\ \hline
      Baseline & 89.4 & 89.3 & \makebox[1.5em]{-} & \makebox[1.5em]{-} & \makebox[1.5em]{-} & \makebox[1.5em]{-} \\ \hline
      Ours w/o ${\cal H}_{\mathcal{T}}$  &88.9  &89.0 & \makebox[1.5em]{-} & \makebox[1.5em]{-} & \makebox[1.5em]{-} & \makebox[1.5em]{-} \\ \hline
      Ours w/o ${\cal H}_{\mathcal{R}}$  & 85.2  & 84.3 & \makebox[1.5em]{-} & \makebox[1.5em]{-} & \makebox[1.5em]{-} & \makebox[1.5em]{-} \\ \hline
      Ours w/o ${\cal H}_{\mathcal{T}},{\cal H}_{\mathcal{R}}$  &84.9  &84.2 & \makebox[1.5em]{-} & \makebox[1.5em]{-} & \makebox[1.5em]{-} & \makebox[1.5em]{-} \\ \hline
      Ours & \textbf{95.1} & \textbf{94.3} & \makebox[1.5em]{-} & \makebox[1.5em]{-} & \makebox[1.5em]{-} & \makebox[1.5em]{-} \\ \hline
      \multicolumn{7}{|c|}{New CAD \cite{Choi:Savarese:Tracking:Collective:Activity:Reog:ECCV2012}} \\ \hline
      UTR-1 \cite{Choi:Savarese:Tracking:Collective:Activity:Reog:ECCV2012,Choi:Collective:Activities:PAMI2014} & 80.8 & 77.0 & 54.3 & 46.3 & \makebox[1.5em]{-} & \makebox[1.5em]{-} \\ \hline
      UTR-2 \cite{Choi:Savarese:Tracking:Collective:Activity:Reog:ECCV2012,Choi:Collective:Activities:PAMI2014} & 83.0 & 79.2 & 53.3 & 43.7 & \makebox[1.5em]{-} & \makebox[1.5em]{-} \\ \hline
      MCTS \cite{AndOrGraph:Scheduling:Activity:ICCV2013} & \makebox[1.5em]{-} & 84.2 & \makebox[1.5em]{-} & \makebox[1.5em]{-} & \makebox[1.5em]{-} & \makebox[1.5em]{-} \\ \hline
      HiRF \cite{HiRF:Group:Activity:ECCV2014} & \makebox[1.5em]{-} & 87.3 & \makebox[1.5em]{-} & \makebox[1.5em]{-} & \makebox[1.5em]{-} & \makebox[1.5em]{-} \\ \hline
      RMI \cite{Recurrent:Modeling:Interaction:Context} & 89.4 & 85.2 & \makebox[1.5em]{-} & \makebox[1.5em]{-} & \makebox[1.5em]{-} & \makebox[1.5em]{-} \\ \hline
      Baseline & 95.3 & 89.0 &70.0  &72.1  &85.7  &92.6  \\ \hline
      Ours w/o ${\cal H}_{\mathcal{T}}$  &\textbf{97.0}  & \textbf{89.3} &76.5  &72.2  &\textbf{90.2}  &\textbf{93.2}  \\ \hline
      Ours w/o ${\cal H}_{\mathcal{R}}$  & 96.8 & 88.0 &74.7  &74.1  &90.0  &93.1  \\ \hline
      Ours w/o ${\cal H}_{\mathcal{T}},{\cal H}_{\mathcal{R}}$  &96.8  &88.0  &73.9  &73.2  &90.0  &\textbf{93.2}  \\ \hline
      Ours & \textbf{97.0} & \textbf{89.3} &\textbf{78.6}  &\textbf{74.7}  &90.0  &\textbf{93.2}  \\ \hline
    \end{tabular}
    \end{minipage}
}
\label{tab1}
\end{table*} 

\begin{table*}[t]
\caption{\em \small Tracking evaluation and comparison with state-of-the-art methods. $\uparrow$ and $\downarrow$ represent ``the-higher-the-better'' and ``the-lower-the-better'', respectively. Bold highlights best results.
\vspace{-0.3cm}
}
\centering
\begin{tabular}{|p{3em}||p{8em}|p{3em}|p{3em}|p{3em}|p{3em}|p{3em}|p{3em}|p{3em}|p{3em}|p{3em}|p{3em}|p{3em}|} \hline
  \makebox[3em]{Dataset} & \makebox[8em]{Method} & \makebox[3em]{Rcll $\uparrow$}& \makebox[3em]{Prcn $\uparrow$} & \makebox[3em]{FAR $\downarrow$} & \makebox[3em]{MT $(\%)\uparrow$}& \makebox[3em]{ML $(\%)\downarrow$} & \makebox[3em]{FP $\downarrow$} & \makebox[3em]{FN $\downarrow$} & \makebox[3em]{IDs $\downarrow$}& \makebox[3em]{FM $\downarrow$} & \makebox[3em]{MOTA $\uparrow$} & \makebox[3em]{MOTP $\uparrow$}\\ \hline
\end{tabular}

\begin{tabular}{|p{3em}||p{8em}|p{3em}|p{3em}|p{3em}|p{3em}|p{3em}|p{3em}|p{3em}|p{3em}|p{3em}|p{3em}|p{3em}|}\hline
  \multirow{8}{*}{\makebox[3em]{CAD \cite{CAD:umich}}}
  & H$^{2}$T \cite{H2T} & \makebox[3em]{\textbf{84.3}} & \makebox[3em]{83.8} & \makebox[3em]{0.90} & \makebox[3em]{74.5} & \makebox[3em]{5.4} & \makebox[3em]{23195} & \makebox[3em]{\textbf{22368}} & \makebox[3em]{474} & \makebox[3em]{722} & \makebox[3em]{67.6} & \makebox[3em]{67.5} \\ \cline{2-13}
  & JPDA \cite{Joint:Probabilistic:Data:Association} &\makebox[3em]{84.1}  &\makebox[3em]{85.5}  &\makebox[3em]{0.79}  &\makebox[3em]{74.0}  &\makebox[3em]{5.4}  &\makebox[3em]{20339}  &\makebox[3em]{22600}  &\makebox[3em]{348}  &\makebox[3em]{901}  &\makebox[3em]{69.6} &\makebox[3em]{63.7}  \\ \cline{2-13}
  & DCEM \cite{Discrete:Continuous} & \makebox[3em]{51.4} & \makebox[3em]{84.3} & \makebox[3em]{\textbf{0.53}} & \makebox[3em]{32.3} & \makebox[3em]{16.2} & \makebox[3em]{13617} & \makebox[3em]{69127} & \makebox[3em]{801} & \makebox[3em]{1025} & \makebox[3em]{41.2} & \makebox[3em]{63.5}  \\ \cline{2-13}
  & POI \cite{POI} & \makebox[3em]{82.3} & \makebox[3em]{76.0} & \makebox[3em]{1.43} & \makebox[3em]{72.1} & \makebox[3em]{5.2} & \makebox[3em]{36944} & \makebox[3em]{25146} & \makebox[3em]{351} & \makebox[3em]{1262} & \makebox[3em]{56.1} & \makebox[3em]{\textbf{67.8}} \\ \cline{2-13}
  & Ours w/o $\mathcal{H}_{\mathcal{T}}$ & \makebox[3em]{84.0} & \makebox[3em]{86.3} & \makebox[3em]{0.74} & \makebox[3em]{74.0} & \makebox[3em]{5.2} & \makebox[3em]{18991} & \makebox[3em]{22717} & \makebox[3em]{355} & \makebox[3em]{623} & \makebox[3em]{70.4} & \makebox[3em]{67.6} \\ \cline{2-13}
  & Ours w/o $\mathcal{H}_{\mathcal{R}}$ & \makebox[3em]{84.0} & \makebox[3em]{85.1} & \makebox[3em]{0.82} & \makebox[3em]{74.0} & \makebox[3em]{\textbf{4.8}} & \makebox[3em]{21002} & \makebox[3em]{22684} & \makebox[3em]{319} & \makebox[3em]{638} & \makebox[3em]{69.0} & \makebox[3em]{67.6} \\ \cline{2-13}
  & Ours w/o $\mathcal{H}_{\mathcal{T}}$,$\mathcal{H}_{\mathcal{R}}$ & \makebox[3em]{84.0} & \makebox[3em]{84.8} & \makebox[3em]{0.83} & \makebox[3em]{\textbf{74.7}} & \makebox[3em]{5.2} & \makebox[3em]{21362} & \makebox[3em]{22717} & \makebox[3em]{360} & \makebox[3em]{630} & \makebox[3em]{68.7} & \makebox[3em]{67.6} \\ \cline{2-13}
  & Ours & \makebox[3em]{84.1} & \makebox[3em]{\textbf{86.6}} & \makebox[3em]{0.72} & \makebox[3em]{74.5} & \makebox[3em]{5.2} & \makebox[3em]{\textbf{18461}} & \makebox[3em]{22647} & \makebox[3em]{\textbf{287}} & \makebox[3em]{\textbf{619}} & \makebox[3em]{\textbf{70.9}} & \makebox[3em]{67.6} \\ \hline
\end{tabular}

\begin{tabular}{|p{3em}||p{8em}|p{3em}|p{3em}|p{3em}|p{3em}|p{3em}|p{3em}|p{3em}|p{3em}|p{3em}|p{3em}|p{3em}|}\hline
  \multirow{8}{*}{}
  & H$^{2}$T \cite{H2T} & \makebox[3em]{87.4} & \makebox[3em]{88.3} & \makebox[3em]{0.37} & \makebox[3em]{81.5} & \makebox[3em]{1.6} & \makebox[3em]{7883} & \makebox[3em]{8572} & \makebox[3em]{117} & \makebox[3em]{232} & \makebox[3em]{75.6} & \makebox[3em]{\textbf{64.7}} \\ \cline{2-13}
  & JPDA \cite{Joint:Probabilistic:Data:Association} & \makebox[3em]{87.6} & \makebox[3em]{88.6} & \makebox[3em]{0.36} & \makebox[3em]{82.6} & \makebox[3em]{1.4} & \makebox[3em]{7660} & \makebox[3em]{8413} & \makebox[3em]{65} & \makebox[3em]{198} & \makebox[3em]{76.3} & \makebox[3em]{62.3} \\ \cline{2-13}
  & DCEM \cite{Discrete:Continuous} & \makebox[3em]{68.5} & \makebox[3em]{88.5} & \makebox[3em]{\textbf{0.28}} & \makebox[3em]{41.9} & \makebox[3em]{7.3} & \makebox[3em]{\textbf{6031}} & \makebox[3em]{21458} & \makebox[3em]{220} & \makebox[3em]{283} & \makebox[3em]{59.3} & \makebox[3em]{62.4} \\ \cline{2-13}
  \makebox[3em]{New} & POI \cite{POI} & \makebox[3em]{87.4} & \makebox[3em]{\textbf{89.0}} & \makebox[3em]{0.34} & \makebox[3em]{82.3} & \makebox[3em]{\textbf{0.8}} & \makebox[3em]{7331} & \makebox[3em]{8594} & \makebox[3em]{\textbf{52}} & \makebox[3em]{271} & \makebox[3em]{76.5} & \makebox[3em]{\textbf{64.7}} \\ \cline{2-13}
  \makebox[3em]{CAD \cite{Choi:Savarese:Tracking:Collective:Activity:Reog:ECCV2012}} & Ours w/o $\mathcal{H}_{\mathcal{T}}$ & \makebox[3em]{87.9}  & \makebox[3em]{\textbf{89.0}} & \makebox[3em]{0.35} & \makebox[3em]{83.9} & \makebox[3em]{\textbf{0.8}} & \makebox[3em]{7411} & \makebox[3em]{8226} & \makebox[3em]{63} & \makebox[3em]{198} & \makebox[3em]{\textbf{76.9}} & \makebox[3em]{\textbf{64.7}} \\ \cline{2-13}
  & Ours w/o $\mathcal{H}_{\mathcal{R}}$ & \makebox[3em]{87.9} & \makebox[3em]{88.9} & \makebox[3em]{0.35} & \makebox[3em]{83.9} & \makebox[3em]{\textbf{0.8}} & \makebox[3em]{7439} & \makebox[3em]{8259} & \makebox[3em]{62} & \makebox[3em]{\textbf{195}} & \makebox[3em]{76.8} & \makebox[3em]{\textbf{64.7}} \\ \cline{2-13}
  & Ours w/o $\mathcal{H}_{\mathcal{T}}$,$\mathcal{H}_{\mathcal{R}}$ & \makebox[3em]{88.1} & \makebox[3em]{88.1} & \makebox[3em]{0.36} & \makebox[3em]{83.9} & \makebox[3em]{\textbf{0.8}} & \makebox[3em]{7726} & \makebox[3em]{\textbf{8096}} & \makebox[3em]{59} & \makebox[3em]{202} & \makebox[3em]{76.7} & \makebox[3em]{\textbf{64.7}} \\ \cline{2-13}
  & Ours & \makebox[3em]{\textbf{88.2}} & \makebox[3em]{88.7} & \makebox[3em]{0.36} & \makebox[3em]{\textbf{84.7}} & \makebox[3em]{\textbf{0.8}} & \makebox[3em]{7630} & \makebox[3em]{8508} & \makebox[3em]{60} & \makebox[3em]{202} & \makebox[3em]{\textbf{76.9}} & \makebox[3em]{\textbf{64.7}} \\ \hline
\end{tabular}

\label{tab2}
\end{table*}




\subsubsection{Implementation}
We implement our method in C++. Experiments are conducted on a machine with a i7-4800MQ CPU (2.8GHz) and 16GB RAM. We use the state-of-the-art person detections \cite{POI} as input, and employ deep re-identification features \cite{POI} as the appearance features for tracking. We set hyperedge degree $m=3$ to balance the performance and speed. The whole pipeline runs in nearly real-time at approximately 20 FPS (not including the detection time). Note that input detectors can be executed in parallel for real-world applications.

\subsubsection{Datasets}
We perform evaluation on three popular collective activity recognition datasets, which are termed CAD \cite{CAD:umich}, Augmented-CAD \cite{Augmented:CAD:umich}, and New-CAD \cite{Choi:Savarese:Tracking:Collective:Activity:Reog:ECCV2012}. Pedestrians in CAD and New-CAD are annotated with target IDs that can be used as ground truth for tracking evaluation.

\textbf{CAD} \cite{CAD:umich} comprises 44 video clips with annotations for $n_C$=$5$ collective activities ({\sc crossing}, {\sc waiting}, {\sc queuing}, {\sc walking}, {\sc talking}), $n_I$=$8$ pairwise interactions: \emph{approaching} (AP), \emph{leaving} (LV), \emph{passing-by} (PB), \emph{facing-each-other} (FE), \emph{walking-side-by-side} (WS), \emph{standing-in-a-row} (SR), \emph{standing-side-by-side} (SS), \emph{no-interaction} (NA), and $n_A$=$2$ individual activities: \emph{standing} and \emph{walking}.

\textbf{Augmented-CAD} \cite{Augmented:CAD:umich} is created by augmenting the CAD dataset. Collective activity {\sc walking} is removed due to its ambiguities in definition, and 2 new collective activities {\sc dancing}, and {\sc jogging} are included. For the newly introduced video clips, there are no annotations for interaction activities, individual activities, nor target identities.

\textbf{New-CAD} \cite{Choi:Savarese:Tracking:Collective:Activity:Reog:ECCV2012} comprises 33 video clips with annotations for $n_C$=$6$ collective activities: {\sc gathering}, {\sc talking}, {\sc dismissal}, {\sc walking-together}, {\sc chasing}, {\sc queuing}, $n_I$=$9$ pairwise interactions: \emph{approaching} (AP), \emph{walking-in-opposite-directions} (WO), \emph{facing-each-other} (FE), \emph{standing-in-a-row} (SR), \emph{walking-side-by-side} (WS), \emph{walking-one-after-the-other} (WR), \emph{running-side-by-side} (RS), \emph{running-one-after-the-other} (RR), \emph{no-interaction} (NA), and $n_A$=$3$ individual activities: \emph{standing}, \emph{walking}, \emph{running}.

\subsubsection{Experimental Setup}
 For evaluating activity recognition, we follow common protocols as in \cite{Choi:Savarese:Tracking:Collective:Activity:Reog:ECCV2012} for CAD and New-CAD, and protocol of \cite{Augmented:CAD:umich} for Augmented-CAD. For evaluating tracking, we ensure fair comparison by running all tracking code using identical input detections.

\subsubsection{Evaluation Metrics}
For activity recognition, we adopt the metrics used in \cite{Choi:Savarese:Tracking:Collective:Activity:Reog:ECCV2012}, \ie, {\em overall classification accuracy} (OCA) and {\em mean-per-class-accuracy} (MCA) as in Table~\ref{tab1}.
Note that the match-error-correction-rate used in \cite{Choi:Savarese:Tracking:Collective:Activity:Reog:ECCV2012} only reflects tracking fragmentation and identity switch. Instead, we adopt the more widely-used CLEAR MOT as tracking metrics to provide further insights for analysis.

\begin{figure*}[t]
\centerline{
  \includegraphics[width=1\linewidth]{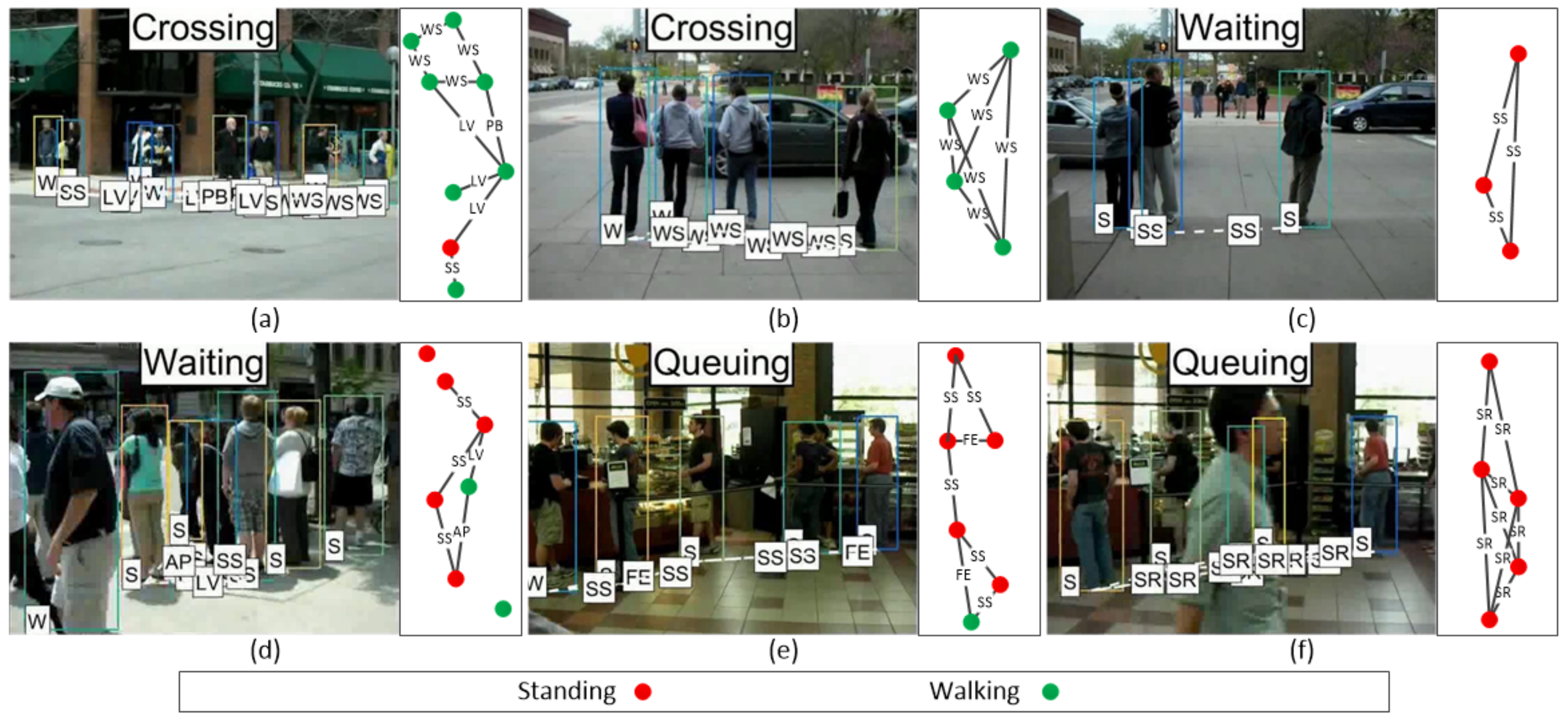}\\ 
}
\caption{\em
Recognized collective activity examples in the CAD dataset \cite{CAD:umich}: (a,b) {\sc crossing}, (c,d) {\sc waiting}, and (e,f) {\sc queuing}. Two individual activities {\em walking} (W) and {\em standing} (S) are detected on each target, while $n_I=8$ possible  pairwise interactions: \emph{approaching} (AP), \emph{leaving} (LV), \emph{passing-by} (PB), \emph{facing-each-other} (FE), \emph{walking-side-by-side} (WS), \emph{standing-in-a-row} (SR), \emph{standing-side-by-side} (SS), \emph{no-interaction} (NA) are detected among the pairs of targets. A top-down view of each scene is illustrated on the right.
}
\label{fig:CAD}
\end{figure*}

\begin{figure*}[t]
\centerline{
  \includegraphics[width=1\linewidth]{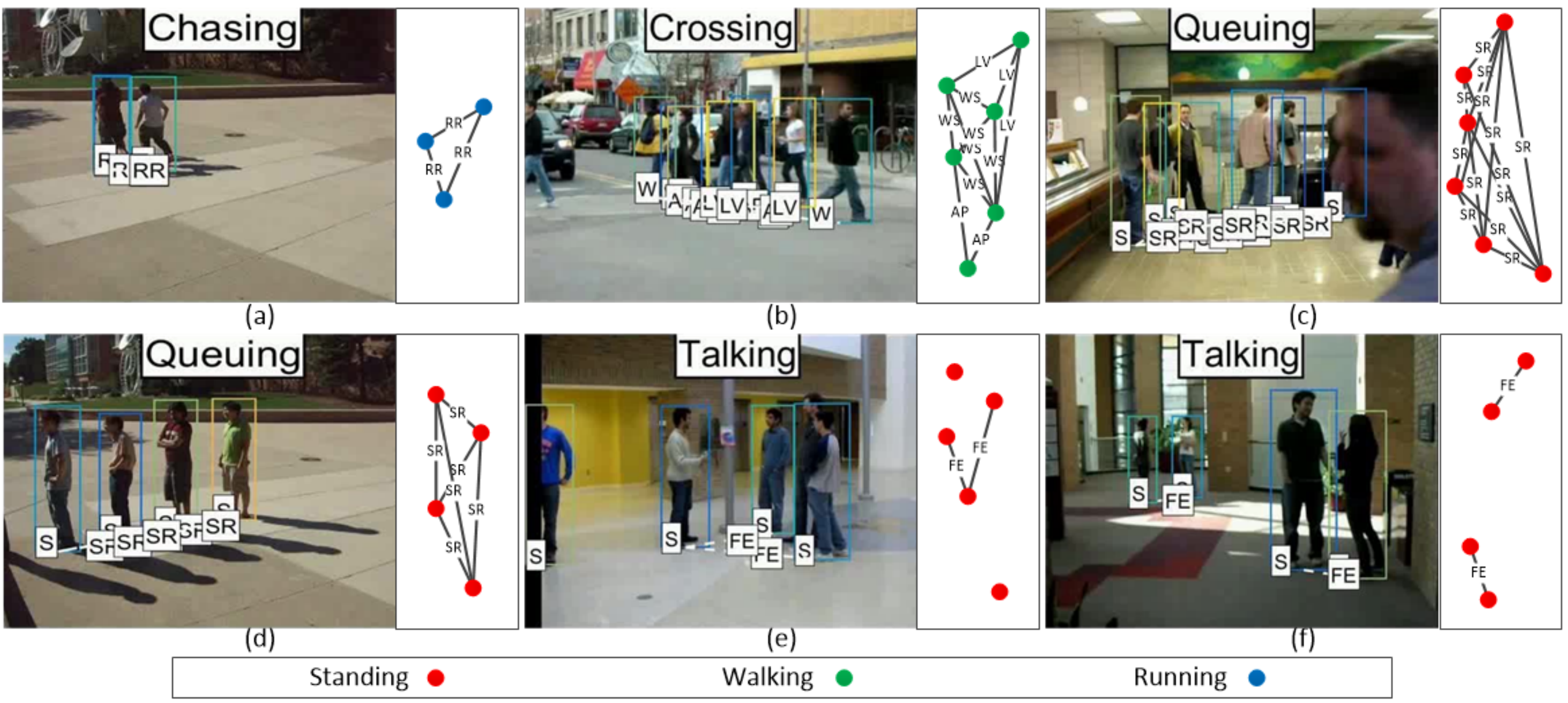}\\ 
}
\caption{\em
Recognized collective activity examples in the NewCAD dataset \cite{Choi:Savarese:Tracking:Collective:Activity:Reog:ECCV2012}:
(a) {\sc chasing}, (b) {\sc crossing}, (c,d) {\sc queuing}, and (e,f) {\sc talking}. Three individual activities   {\em walking} (W), {\em standing} (S), and running (R) are detected on each target, while $n_I=9$ possible pairwise interactions: \emph{approaching} (AP), \emph{walking-in-opposite-directions} (WO), \emph{facing-each-other} (FE), \emph{standing-in-a-row} (SR), \emph{walking-side-by-side} (WS), \emph{walking-one-after-the-other} (WR), \emph{running-side-by-side} (RS), \emph{running-one-after-the-other} (RR), \emph{no-interaction} (NA) are detected among the pairs of targets. A top-down view of each scene is illustrated on the right.
}
\label{fig:newCAD}
\end{figure*}

\subsubsection{Compared Methods}

We compare our method with 13 state-of-the-art activity recognition methods \cite{HiRF:Group:Activity:ECCV2014,AndOrGraph:Scheduling:Activity:ICCV2013,Learning:Latent:Constituents,Choi:Savarese:Tracking:Collective:Activity:Reog:ECCV2012,
Choi:Collective:Activities:PAMI2014,Augmented:CAD:umich,Structure:Inference:Machines,Deep:Structured:Models,Visual:recognition:by:counting:instances,
Hierarchical:Deep:Temporal:Model,Combining:Per-frame:and:Per-track:Cues,Flow:Model,Recurrent:Modeling:Interaction:Context} and a few baseline methods created by ourselves. These baseline methods accept the tracking results of \cite{H2T} as input, and recognize activities using only our probabilistic rules {\em e.g.}, Eqs.(\ref{eq:individual:correlation:measure}, \ref{eq:interaction:prob}, \ref{eq:collective:prob}) with details in Table~\ref{tab:interaction:formulas}, but not the complete staged hypergraph optimizers. For tracking evaluation, we compare against 4 state-of-the-art trackers \cite{Discrete:Continuous,Joint:Probabilistic:Data:Association,H2T,POI} with available code. We also develop several variants of our method by replacing the hypergraph formulations with the ordinary graph formulations. This justifies the effect of hypergraph formulations {\em w.r.t.} performance in both activity recognition and tracking.

\subsection{Results and Analysis}

\begin{figure*}[t]
\centerline{
  \includegraphics[width=0.95\linewidth]{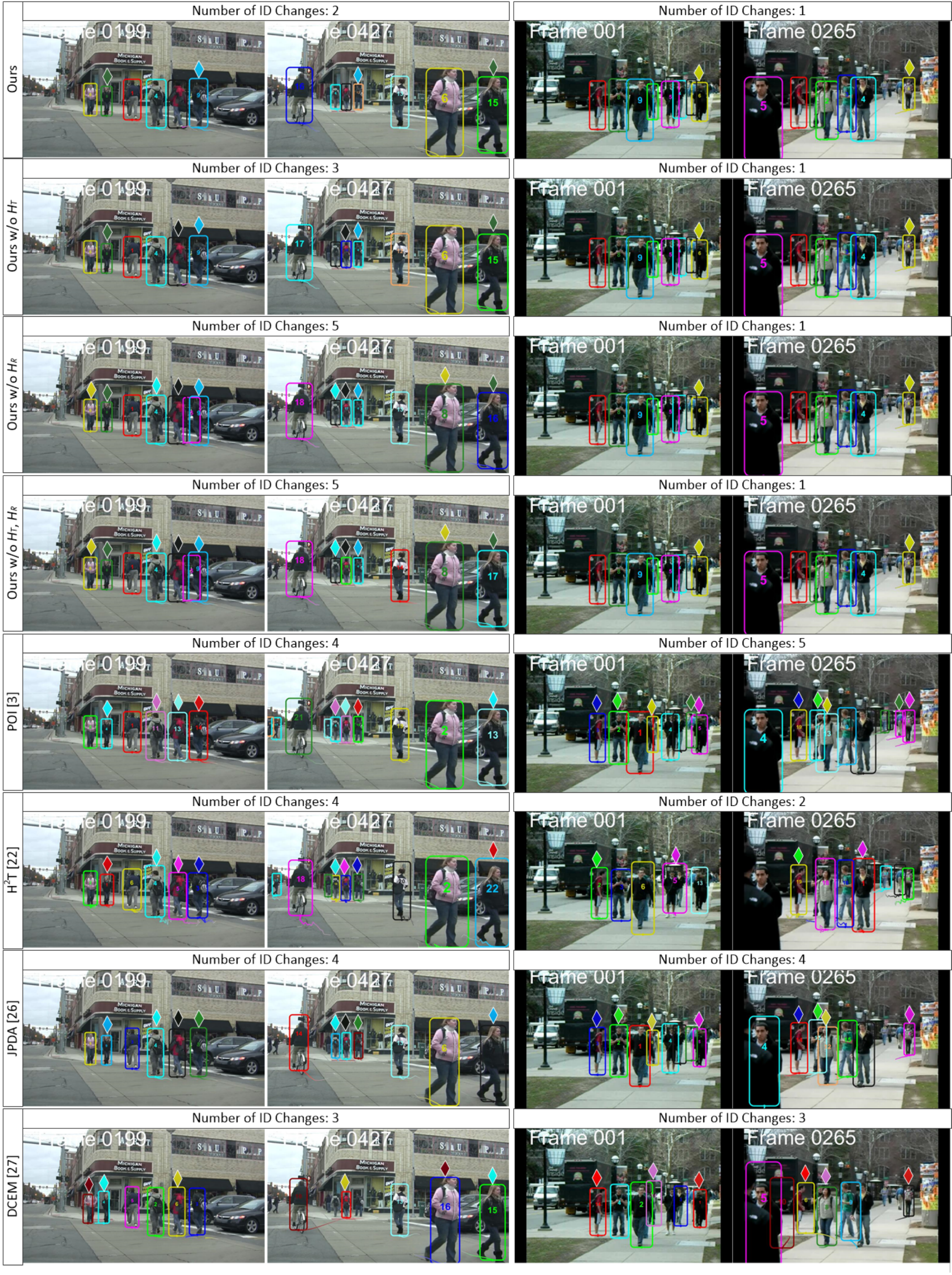}\\ 
}
\caption{\em
Example tracking results. For each sequence, we show two frames across a long period of time, where the subjects have experienced heavy occlusions. The number of target ID changes are shown in each experiment. Colorful diamonds visualize the corresponding target ID changes.
}
\label{fig:tracking}
\end{figure*}

\subsubsection{Activity Recognition}

Table~\ref{tab1} shows the evaluation results in terms of accuracy of our method and other state-of-the-art methods on the three activity recognition datasets. Part of our recognition results are shown in Fig.~\ref{fig:CAD} and Fig.~\ref{fig:newCAD}. The efficacy of our method for individual activity recognition is demonstrated by the high accuracy score of approximately $90\%$. Our method outperforms the state-of-the-art methods by a noticeable margin for collective activity recognition. The performance of our method is significantly better than \cite{Choi:Savarese:Tracking:Collective:Activity:Reog:ECCV2012,Choi:Collective:Activities:PAMI2014}, which are, to the best of our knowledge, the only works that evaluate pairwise interactions. Comparison with the three variants of our method shows that both hypergraphs $\mathcal{H}_{\mathcal{T}}$ ($\S$~\ref{sec:stage1}) and $\mathcal{H}_{\mathcal{R}}$ ($\S$~\ref{sec:stage2}) contribute to the improvements in the collective and interaction activity recognition.
The experiments show that $\mathcal{H}_{\mathcal{R}}$ is more influential than $\mathcal{H}_{\mathcal{T}}$, which is expected, since the main purpose of $\mathcal{H}_{\mathcal{T}}$ is to improve tracking and serve as a base for activity recognition. We also notice positive correlations between the recognition accuracy of $C$ and $I$.
Overall, {\sc walking} is difficult to recognize due to its ambiguity {\em w.r.t.} {\sc crossing}. Nonetheless our method can still recognize it well due to our probabilistic formulation.

\subsubsection{Tracking}

Table~\ref{tab2} shows the comparison of our method and other state-of-the-art tracking methods on the CAD and New-CAD. On the CAD dataset, our method achieves the best performance in most measures, \eg, MOTA, fragmentation (FM), identity switch (IDs), false positives (FP), mostly tracked targets (MT), precision (Prcn), and recall (Rcll). This is due to the incorporation of the high-order correlations in $\mathcal{H}_{\mathcal{T}}$ and $\mathcal{H}_{\mathcal{R}}$. Specifically, we use $\mathcal{H}_{\mathcal{T}}$ to model high-order correlations of individual activities, which improves the tracklet association. The use of $\mathcal{H}_{\mathcal{R}}$ to model high-order correlations of interaction activities improves occlusion recovery.
This is further confirmed that after replacing $\mathcal{H}_{\mathcal{T}}$ or $\mathcal{H}_{\mathcal{R}}$ with ordinary graphs, the tracking performance decreases in most measures.
New-CAD is less challenging than CAD in terms of tracking, because there are fewer occlusions and crossing occasions. Thus many compared methods yield performances closer toward saturation. However, our method still achieves the highest score in several measures, \ie, MOTA, mostly tracked targets (MT), and recall (Rcll). Both \cite{H2T} and our method rely on the cohesive cluster search on the hypergraph, but our method consistently outperforms \cite{H2T} by a significant margin because of {\em (i)} the modeling of high-order correlations of individual activities, and {\em (ii)} successful occlusion recovery. We visualize the tracking result comparisons for several sequences in Fig.~\ref{fig:tracking}. For each sequence, we show two frames across a along period of time, where the subjects have experienced heavy occlusions. It is clear that our method (especially the one with hypergraph optimizers) produces the fewest ID changes, thus it is more robust than the competing methods for occlusion handling.

\section{Conclusion}

We present a novel multi-stage framework for solving the joint tasks of multi-person tracking and group activity recognition. This approach can effectively address not only the {\bf where} and {\bf when} problem by visual tracking, but also the {\bf who} and {\bf what} problem by  recognition. By explicit modeling of correlations among individual activities, pairwise interactions, and collective activities using hypergraphs, we can effectively improve recognition and tracking. Our method can track targets with occlusion recovery, identify correlated pairwise interactions, and recognize group collective activities. Experimental evaluations demonstrate that our method outperforms state-of-the-art methods in both tasks of tracking and activity recognition. Our method runs in nearly real-time (not counting input detections), and is applicable to a variety of real-world applications including video surveillance and situational awareness. Implementation code will be released upon the publication of this work.



\bibliographystyle{ieeetran}
\bibliography{reference}

\end{document}